# Socially Consistent Multi-Robot Navigation Using Decoupled Planning and Trajectory Coordination

Matthew M. Sato* and Kincho H. Law
Engineering Informatics Group, Department of Civil and Environmental Engineering
Stanford University, Stanford, CA 94305

## Abstract

The successful integration of mobile robots in human-centric environments requires navigation that is not only safe and efficient, but also consistent and compliant with social conventions—key precursors for human comfort and acceptance. Most human-aware navigation research targets the local planner, generating short-horizon, sensing-dependent reactive behavior around nearby humans. However, some social conventions are long-horizon routing choices that are better established at the global planning level, where the positions of individual humans are either unobservable or likely to be stale by the time the robot arrives. We develop a partially decentralized global planning system that encodes social conventions and coordinates multiple robots, complementing local planners that react to dynamically sensed humans. First, we introduce a general mechanism for embedding sensing-independent social conventions into a modified A* cost function. Planned paths are shared across the fleet to collaboratively build a shared social graph of established routes, enforcing path consistency and reducing future planning effort. Second, we leverage the resulting structure of these socially consistent paths to formulate multi-robot trajectory coordination as a mixed-integer convex program. While the underlying optimization is solved centrally, the constraint generation is distributed across the fleet, enabling efficient computation of conflict-free trajectories. Through simulation and hardware experiments, we demonstrate that viewing global social path planning and multi-robot trajectory coordination as a single system, rather than independent problems, produces socially consistent, repeatable paths and simplifies multi-robot coordination.

## 1. Introduction

The rapid improvements in artificial intelligence and robotics hardware enable the deployment of autonomous mobile robots in increasingly complex and human-centric environments, including restaurants, warehouses, and airports. To facilitate and encourage widespread adoption, mobile robots should be seamless, safe, and accepted by humans. Human acceptance of mobile robots is highly dependent on the robot's motion, with humans concerned with both the actual and perceived level of comfort and safety around mobile robots [1,2]. Mobile robot motion that is aggressive or unpredictable can therefore delay or prevent adoption of mobile robots in human environments due to human discomfort [3]. Developing motion planning that accounts for human concerns is therefore a critical prerequisite for the widespread adoption of mobile robots in human spaces.

Mobile robot navigation is typically decomposed into a *global planner*, which computes a route across the full environment before motion begins, and a *local planner*, which reactively adjusts robot trajectory based on current sensor data. The separation of global and local planning mirrors how humans may navigate: a person forms a long-term route to a destination without knowing the precise position of everyone they may encounter, then reacts and steps around people they encounter. Because global planning considers regions of the environment outside the robot's current sensing range, robots cannot reliably know where humans will be in the future, and any such knowledge would likely be stale by the time the robot arrived. Consequently, the global planner should not route around specific, instantaneous human positions. Instead, local planners exist to dynamically and socially avoid humans, whether through purely reactive methods [4] or methods with a limited lookahead horizon [5–7]. Long-horizon route consistency, chosen independently of current human occupancy, is instead a property of the global planner and the focus of this work. As illustrated in Figure 1, a socially compliant global path may deviate from the shortest path to conform to social conventions. We emphasize that our proposed global planner is a complement, not a replacement, to social local planners: the global planner supplies long-horizon socially compliant waypoints, while instantaneous obstacle avoidance around dynamic humans remains the responsibility of the local planner.

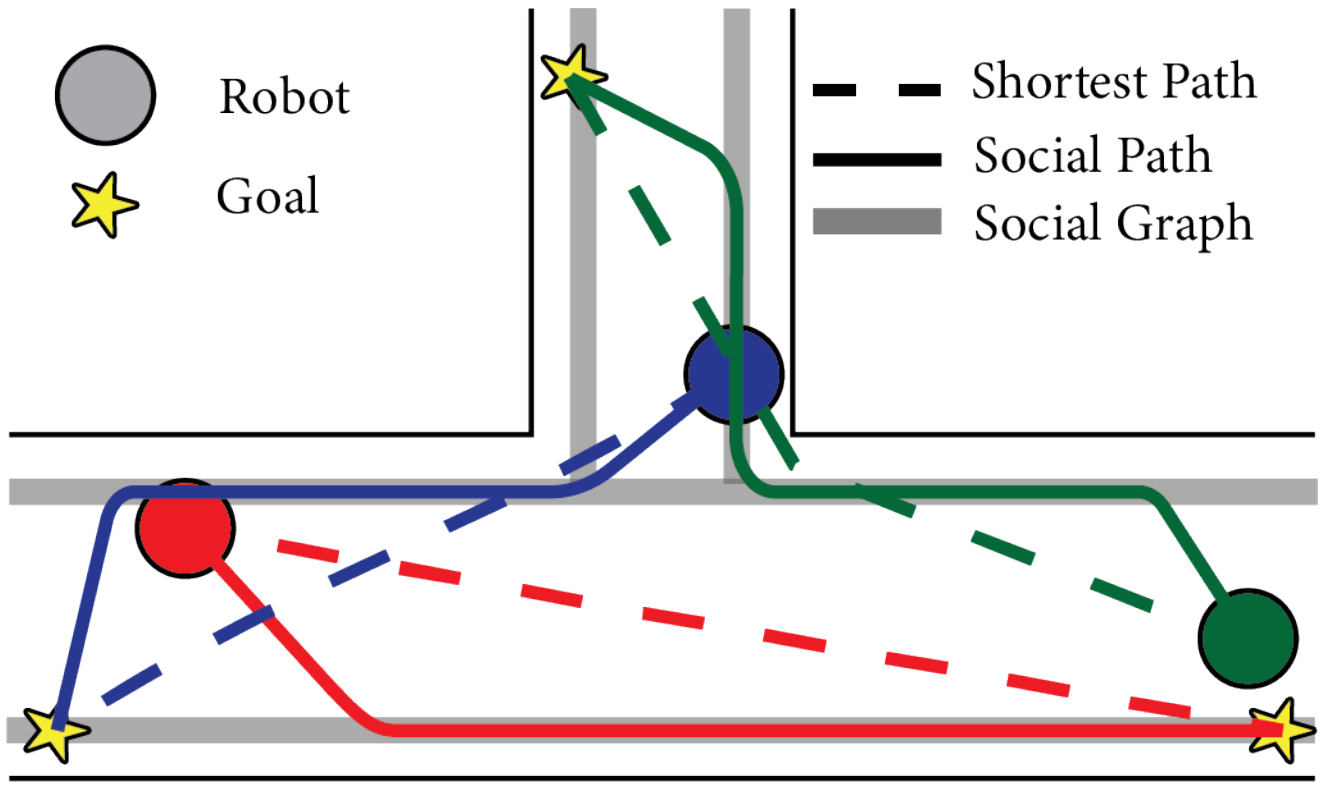

**Fig. 1.** Illustration of global social path planning for a multi-robot fleet. Rather than following the shortest path, each mobile robot follows a social path that trades path length for conformance to social conventions. Robot paths are consolidated into a shared social graph which enforces consistency across the fleet and simplifies multi-robot trajectory coordination.

Prior work explores social conventions in navigation that are conditioned on sensed humans at planning time and exhibits global-planning-like features, for example by planning routes over medium-scale distances [8–10]. Beyond spatial extent, what distinguishes these extended-horizon approaches from true global planning is their dependency on sensed human positions when computing social cost: each replans as the robot advances and new observations arrive, rather than committing to a route in advance. In [10], a social cost layer over a 16 m × 16 m region centered on the robot is updated continuously as the robot moves or as humans are newly sensed. In [8,9], plans are formed relative to currently observed human poses and re-executed whenever new sensing invalidates the current plan. Existing taxonomies of social-aware navigation architecture divide navigation into a global planner that uses social costs derived from a static map, and a local planner informed by sensor-driven human detection, tracking, and prediction [11]. By this taxonomy, [8–10] fall within the local planner branch despite their extended spatial reach, due to their dependence on sensed human states.

We provide a general, extensible mechanism for embedding sensing-independent social cost functions into a global A* planner, producing paths we term socially compliant. We clarify that our use of “global” refers to path planning across the full geometry of the environment, not to any claim that the social norms embedded in the planner are themselves universal. We combine the global planner with a fleet-wide method that consolidates the routes over time into a shared social graph, a set of common paths planned across the fleet, allowing the robots to achieve social consistency and reducing future planning effort. As we describe next, a coordination method then exploits this graph structure to make multi-robot trajectory coordination tractable. We implement a basic social cost function (a preference for right-hand hallway travel and minimum wall clearance) intended to demonstrate the framework and its downstream trajectory coordination benefits, not to constitute a comprehensive model of human social navigation. Richer social cost functions can be substituted into the same framework without altering the global planning or coordination methodology, provided they remain sensing-independent at the global stage. We note that such cost functions may still draw on sensed or historical information, for example differing social patterns throughout the day, but should remain fixed over the course of a robot's traversal, rather than updating in response to sensor data gathered by the robot while moving.

A social cost function shapes individual paths but doesn't inherently lead independently planned robot paths to converge on the same routes. Even if robot paths converge to a common set of paths, the structure of common paths alone does not resolve collisions between multiple robots. Extending social compliance and consistency to a coordinated, multi-robot fleet is a separately difficult, combinatorial problem whose solution space grows exponentially with fleet size. Existing taxonomies of socially aware navigation, including the frameworks surveyed in [11], are formulated for single robots. Although a growing body of work has begun to address social conventions in multi-robot navigation, this work again occurs primarily at the local planning level [12–15]. Meanwhile, existing global multi-robot coordination methods operate independently of social structure and treat every potential conflict as a distinct case to resolve [4,16,17]. We observe that once individual robot paths are restricted to a shared set of social behaviors, the space of possible conflicts collapses to a much smaller number of well-defined collision scenarios, which we exploit to formulate trajectory coordination as a mixed-integer convex program.

We formulate trajectory coordination as a mixed-integer convex program that assigns collision-free arrival times at a small set of conflict waypoints identified from the social path structure. Because the number of optimization variables scales with the number of possible conflicts rather than the number of robots, path length, or path resolution, the convex program scales more favorably than approaches that do not exploit social structure. Constraint generation can further be distributed across the robot fleet, enabling efficient problem generation; however, the resulting convex program itself is solved centrally. We evaluate the coordination formulation at moderate fleet sizes characteristic of human-centric deployments, rather than for swarm-scale fleets, and validate the approach on a small hardware fleet.

As shown in Figure 1, when multiple robots' socially compliant paths converge into a smaller set of shared route segments, they form a structure our coordination formulation exploits to reduce potential conflicts. Together, these three components—a sensing-independent cost function, a shared route structure, and a trajectory coordination method—form a system for global social planning of mobile robot fleets. The result is a set of long-horizon routes that complement existing social local planners and are structured specifically to simplify multi-robot coordination. While each of these ideas has been introduced separately in prior work, to the best of our knowledge they have not previously been combined to form a global social multi-robot planner. Specifically, this paper contributes a system containing:

- A framework for embedding sensing-independent social cost functions into global path planning.
- A mechanism for consolidating socially compliant routes across a fleet into a shared social graph, achieving path consistency without requiring learned models.
- A mixed integer convex formulation of multi-robot trajectory coordination that exploits the social route structure to reduce possible collision scenarios, evaluated at moderate fleet scale in simulation and validated on hardware.

The remainder of this paper is organized as follows: Section 2 reviews related works, Section 3 details the proposed methodology for path planning and control, Section 4 presents the results from simulated mobile robot experiments and demonstrations on hardware, and Section 5 contains concluding remarks.

## 2. Related Work

Mobile robot path planning is a foundational problem in robotics, with extensive research dedicated to finding computationally efficient and optimal paths. Classic algorithms such as A* [18], Rapidly-exploring Random Trees (RRT*) [19], and Probabilistic Roadmaps (PRM) [20] provide the bedrock for many modern motion planners. These algorithms excel at finding optimal paths according to a defined cost function, which is typically path length or travel time. However, these algorithms do not inherently consider the social context of navigating in human-centric environments or provide the means for determining multi-robot trajectories. This section reviews five areas of related research that are relevant to the methods presented in this work: multi-robot coordination, local social navigation for single robots, extended-horizon social navigation, learning-based approaches to social navigation, and multi-robot social navigation.

### 2.1 Multi-Robot Path Planning and Coordination

Coordinating the motion of multiple mobile robots, termed multi-agent pathfinding (MAPF), is a computationally challenging problem. MAPF methods are broadly categorized into centralized and decentralized approaches. Centralized planners aim to find the globally optimal solution for the entire fleet. Conflict-based search (CBS) [16] and its variants are prominent optimal solvers that first plan individual paths and then resolve collisions with a search tree. However, the computational complexity of CBS grows large with the number of conflicts. Priority motion planning [17] offers a computationally simpler alternative by assigning a fixed priority to robots and planning paths sequentially. However, priority planning can lead to suboptimal solutions where low-priority robots take excessively long paths. Meanwhile, decentralized planners focus on local interaction for scalability. Optimal reciprocal collision avoidance (ORCA) [4] and its non-holonomic extension (NH-ORCA) [21] are state-of-the-art reactive methods. Each agent independently computes collision-free velocities based on the current state of its neighbors. ORCA is scalable and effective at preventing immediate collisions. Both the centralized and decentralized approaches are effective at solving the multi-robot coordination problem, but ignore social conventions, which may lead to undesirable behavior in human-centric environments.

### 2.2 Local Social Navigation for Single Robots

A large body of research has focused on socially aware single robot navigation at the local planning level. Much of the socially aware mobile robot work is based on proxemics [22], the sociological study of human spatial behavior and the rules that define the personal spaces of humans. A common approach is the social force model [23], which models humans as particles subject to attractive and repulsive forces. Planners can incorporate the social force model directly or to generate a social cost map that penalizes moving too close to humans [24–26]. Other methods use sensor data, such as from fixed external cameras, to generate a heatmap of human activity and direct mobile robots to less crowded areas [27]. Other single-robot approaches modify local trajectory optimizers to incorporate social rules. Methods based on the timed elastic band (TEB) algorithm dynamically adjust mobile robot trajectories to maintain appropriate distances from humans, anticipate their movements, and adhere to social norms [5–7]. These methods are effective at shaping instantaneous, ego-centric robot behavior, but by design they operate over the robot's immediate sensing horizon and do not establish consistency over a complete route or across a fleet.

### 2.3 Extended-Horizon Social Navigation

Other bodies of work extend social cost reasoning beyond a robot's immediate horizon, computing a full human-aware trajectory within the robot's observed surroundings [8], or exploiting space expected to become free as a human moves [9]. More recently, a general social cost layer approach [10] uses a neural network to predict a social cost map over a fixed local region, which is integrated as an additional cost layer for a grid-based planner that then plans a complete path using this cost map. While these approaches consider a broader spatial context than purely reactive local planners, each remains fundamentally dependent on the sensed positions of specific humans at planning time, and each approach replans as the robot advances and new observations arrive. The frameworks in [11] separate socially-aware navigation architectures into those derived from a static map (global planner) and those informed by human detection, tracking, and prediction (local planner). By the taxonomy in [11], the approaches above are best characterized as extended-horizon local planners: although they reason over a larger spatial context than a purely reactive planner, they remain dependent on real-time human sensing. Our approach instead computes social costs from the static map alone, complementing the prior work in this section by providing long-horizon, sensing-independent waypoints. Further, these extended-horizon approaches do not consider multi-robot settings, a gap our global approach addresses directly.

### 2.4 Learning-Based Social Navigation

To move beyond hand-crafted social rules, researchers have turned to reinforcement learning (RL). Inverse RL learns a social cost function from human demonstrations, allowing mobile robots to mimic human behaviors while path planning [28]. Other approaches use deep RL to guide mobile robot navigation. Approaches such as socially aware collision avoidance with deep reinforcement learning (SA-CADRL) [29] and crowd-aware navigation with deep reinforcement learning [30] train a deep neural network to serve as the control policy, mapping sensor inputs to actions. Deep RL can produce sophisticated behaviors learned from simulation or real-world data. A comprehensive review of deep RL techniques for social navigation is found in [31]. However, learning-based methods have limitations. Namely, the black-box nature of RL approaches can make mobile robot behavior difficult to predict or guarantee, which is counter to the goals of this work. Furthermore, learning-based approaches typically focus on ego-centric local interactions and do not enforce the global consistency that is an objective of this work.

### 2.5 Multi-Robot Social Navigation

Recent work has begun to address the multi-robot social navigation scenario in human environments, extending social navigation from single robots to environments with both robot-robot and robot-human interactions. Most commonly, these approaches use learning-based policies with centralized training and decentralized execution. For example, MSA3C [12] uses a learned temporal-spatial graph social encoder to determine robot motion based on observed humans and robots within the field of view. SAMARL [13] uses a spatial-temporal transformer with an interaction graph to encode dependencies between robots and humans and to provide realistic robot dynamics. MultiSoc [14] uses two graph neural networks to determine influence between agents, and an RL approach to learn the multi-agent policy according to these interactions. Complementing these graph and attention approaches, CEMRRL [15] provides an RL approach to improve coordinated exploration in social multi-robot tasks, aiming to improve learning efficiency.

A recurring limitation of these learning-based single- and multi-robot social navigation approaches is that social behaviors emerge from reward shaping and the resulting policy, rather than being directly controlled at the planning level. The end-to-end policies simultaneously govern the high-level route, local collision avoidance, and inter-robot coordination. These single coupled policies have no explicit mechanism for enforcing map-level social conventions across long horizons or exploiting the structural regularities of socially consistent routes to coordinate mobile robots. Taken together, the existing work leaves a gap that motivates our approach. Multi-robot coordination methods are effective, but either ignore social conventions entirely (Section 2.1) or focus only on the local planning level (Section 2.5). Single-robot social navigation (Sections 2.2-2.4) focuses on sensing-dependent local planners and is not designed to enforce fleet-wide consistency or coordinate multiple robots. No existing method embeds social norms into a sensing-independent global planner, enforces fleet-wide consistency, and exploits the resulting route structure to make multi-robot trajectory coordination tractable.

## 3. Materials and Methods

This section details our proposed framework for socially consistent multi-robot navigation and trajectory coordination. Our approach decouples the complex navigation task into distinct, sequential stages as shown in Figure 2. In Section 3.2, we describe the first stage, in which each mobile robot independently plans a socially compliant geometric path using a modified A* planner and a shared social graph to ensure fleet-wide consistency. In Section 3.3, we detail the second stage, in which mobile robots coordinate trajectory arrival times along their planned paths using a mixed-integer convex program (MICP).

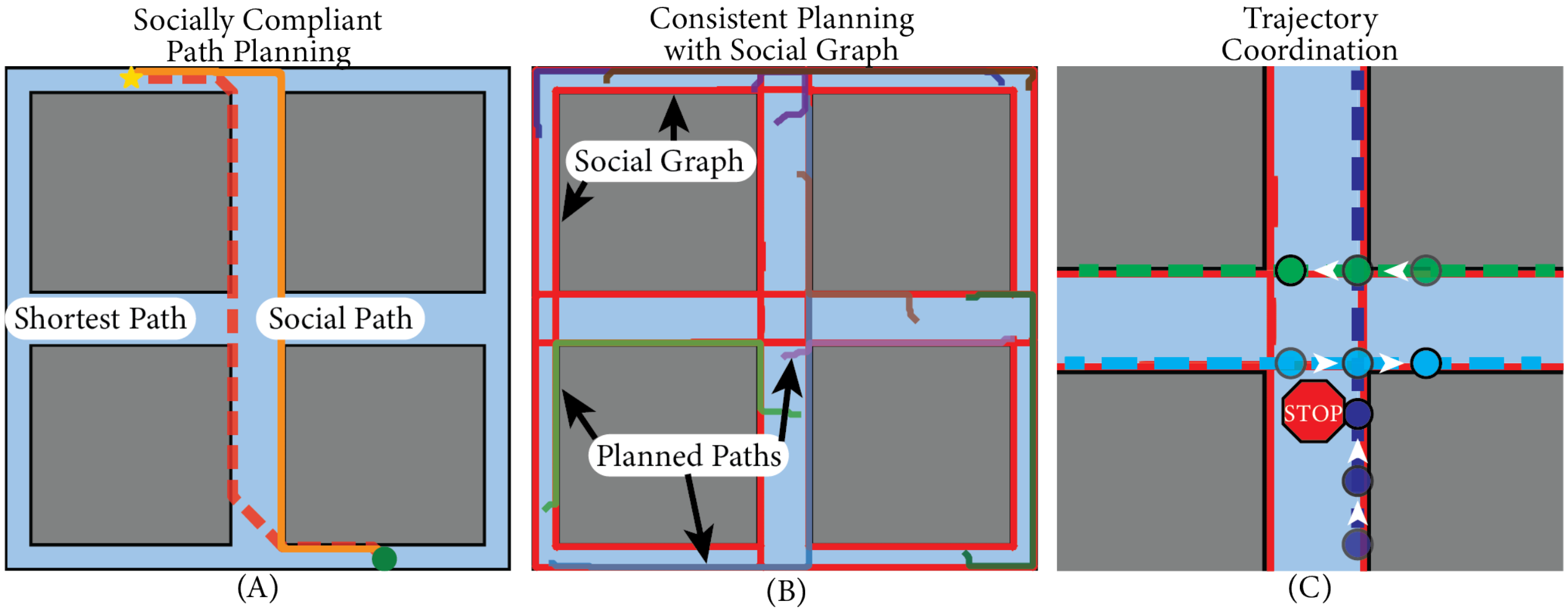

**Fig. 2.** Overview of the decoupled navigation framework. (A) Social A* planner generates a socially compliant geometric path. (B) Multiple paths are aggregated to form a shared social graph. (C) A convex optimization solver coordinates the trajectories of mobile robots to prevent collisions.

### 3.1 Preliminaries: Map Representation and Planning Space

Our framework computes long-horizon global paths and therefore assumes a known, static 2D environment, represented as an occupancy grid in which space is discretized into cells labeled as occupied, free, or unknown. We assume the map has been constructed a priori using a mapping algorithm such as simultaneous localization and mapping (SLAM) [32]. Consequently, the social costs used by our global planner (Section 3.2.1) are computed once from this static map and do not depend on active human sensing at planning time. For dynamic objects such as humans, local approaches such as those introduced in Section 2 can be used with our approach to provide a complete navigation solution. To simplify the pathfinding problem, we transform the occupancy grid into configuration space (C-space) [33]. In C-space, obstacles are inflated by the mobile robot's radius, allowing the planner to model the mobile robots as point masses and simplifying the navigation problem.

## 3.2 Global Path Generation

The first stage of our framework determines a feasible and socially aware geometric path for each mobile robot. While each mobile robot plans independently, the fleet achieves emergent cooperation through two mechanisms detailed in this section: a common social cost function applied by every robot (Section 3.2.1) and a collaboratively built social graph that enforces consistency across mobile robots (Section 3.2.2). As introduced in Section 1, a path is *socially compliant* if it adheres to the social cost function, and the fleet is *socially consistent* when robots converge on the same established socially compliant routes, which independent compliance does not guarantee.

### 3.2.1 Socially Compliant Path Planning for a Single Robot

The core of our single robot path planner is a modified A* search algorithm [18] operating on the C-space grid. We use the A* search algorithm because of its deterministic behavior, optimality guarantees, and the flexibility its cost function structure affords for encoding social norms. The A* cost function is modified to shift from finding the shortest path to finding a path that balances path length with adherence to social norms.

As summarized in Algorithm 1, A* finds a path by iteratively expanding the node $x$ which minimizes the function $f(x) = g(x) + h(x)$, where $g(x)$ is the accumulated cost from the start node to $x$, and $h(x)$ is a heuristic estimate of the cost from $x$ to the goal. We use the Manhattan distance as the heuristic:

$$h(x) = \lVert x_{\text{goal}} - x \rVert_1 \tag{1}$$

which is consistent and admissible, guaranteeing that A* finds the lowest cost path based on the cost function $g(\cdot)$ [18].

**Algorithm 1:** Pseudocode for the A* pathfinding algorithm.

```
initialize priority queue open_set and empty set closed_set
add x_init to open_set, initialize g[x_init] := 0
while open_set is not empty:
    x_current := x with smallest priority, p, from open_set
    if x_current = x_goal → return
    move x_current from open_set to closed_set
    for each x_neigh in neighborhood of x_current:
        if x_neigh in closed_set → continue
        new_g := g[x_current] + c[x_current, x_neigh]
        add x_neigh to open_set with priority new_g + h[x_neigh]
```

We embed social norms through the definition of the edge cost, $c(x', x)$, which is the cost the mobile robot incurs when transitioning from $x'$ to $x$. The cost function is $g(x) = g(x') + c(x', x)$: the cost of the previously visited node $x'$ and the edge cost to reach $x$ from $x'$. To enforce social behavior while also achieving short path distances, we define the edge cost as a combination of Euclidean distance and a social cost, $\phi(x', x)$:

$$c(x', x) = \lVert x - x' \rVert_2 + \phi(x', x) \tag{2}$$

The added social cost allows any social convention that can be encoded as $\phi(x', x)$ to be incorporated directly into the A* edge cost, without further modification to the search algorithm itself. The social cost $\phi$ translates abstract social rules into a numerical penalty and is therefore a design decision independent of the planner, ranging from simple geometric heuristics we use here to more sophisticated or context-dependent social cost models. For example, the social cost could depend on historically sensed information or use a learned cost function. To demonstrate the social cost mechanism, we implement a simple instantiation of $\phi$ using three social components that follow social convention in the United States:

$$\phi(x', x) = w_{\text{right}} C_{\text{right}}(x', x) + w_{\text{const}} C_{\text{const}}(x', x) + w_{\text{buf}} C_{\text{buf}}(x', x) \tag{3}$$

where:

- $C_{\text{right}}(x', x)$ enforces a keep-right behavior. The cost can be defined as the distance to the nearest obstacle on the right, $d_{\text{right}}(x)$. For example, $C_{\text{right}}(x', x) = d_{\text{right}}(x)$ penalizes any path far from the right wall.
- $C_{\text{const}}(x', x)$ promotes smooth, parallel motion by penalizing changes in the distance to the right wall between steps: $C_{\text{const}}(x', x) = d_{\text{right}}(x) - d_{\text{right}}(x')$.
- $C_{\text{buf}}(x', x)$ is active in wide-open areas (where $d_{\text{right}}(x)$ exceeds a threshold). $C_{\text{buf}}(x', x)$ encourages leaving a buffer for oncoming traffic by penalizing proximity to the left wall: $C_{\text{buf}}(x', x) = \max(0, d_{\text{buf}} - d_{\text{left}}(x))$, where $d_{\text{buf}}$ is the desired buffer width.

The non-negative weights $w_{\text{right}}$, $w_{\text{const}}$, and $w_{\text{buf}}$ are tuned for different environments and social conventions. The social cost weights are set to zero for all nodes within a certain radius of the start and goal positions, which prevents penalizing necessary maneuvers at the beginning or end of a path. The performance of the social planner depends on the selection of the weights and threshold in the cost function. While these parameters can be learned, for our work we select them as follows: $w_{\text{right}}$ and $w_{\text{buf}}$ are weighted highest to establish strong lane discipline, followed by $w_{\text{const}}$ to prevent jagged paths. Last, $d_{\text{buf}}$ should be larger than the robot diameter to ensure adequate passing clearance. We emphasize that these social cost functions are just one set of possible cost functions, and any cost function can be substituted to achieve the desired social behavior, provided it remains sensing-independent (computed once from a static map rather than updated in response to actively sensed humans during robot motion).

### 3.2.2 A Social Graph for Fleet-Wide Path Consistency

The single-robot planner from Section 3.2.1 encourages each path to be socially compliant, but the planner has two main limitations when operating in a multi-robot setting: a large computational expense when computing the social cost $\phi$ and no guarantee of social consistency across mobile robots. To address both issues, we introduce a mechanism for the robot fleet to collaboratively build a directed subgraph, hereafter the *social graph*. The social graph represents the established, socially compliant routes in the environment. Planning over the much smaller social graph is significantly faster and enforces consistent path choices across the entire fleet.

The social graph, $H(V_H, E_H)$, is built and refined dynamically during robot operations through a three-step process:

1) **Path Aggregation**: Every successfully planned path is broadcast to other mobile robots. Each mobile robot maintains a set of eight directional heatmaps, one for each cardinal and intercardinal grid direction (north, northeast, west, etc.). When a path is received, the corresponding heatmap cell is incremented for each segment of the path. The distinction between directions distinguishes between traffic moving in opposite directions within the same hallway, preventing the creation of a single lane highway.

2) **Subgraph Extraction**: Periodically, the directional heatmaps are processed to form the social graph. A directed edge *(x', x)* is added to the social graph, *H*, if *x'* and *x* are adjacent and the value in the corresponding directional heatmap exceeds a predefined threshold. Edges added to the social graph are given edge weights equal to the Euclidean distance between the two nodes, $\|x - x'\|_2$.
3) **Graph Pruning**: To filter noise from transient or exploratory paths, the extracted social graph is pruned. The social graph's weakly connected components are computed, and any components with a node count below a minimum node size threshold are discarded. The pruning step ensures *H* contains only well-established, significant trajectories, improving robustness to one-off path deviations.

Once computed, the social graph *H* biases future path planning. The A* global planner's edge cost function is modified to favor paths along the social graph, with the new edge function:

$$c'(x',x) = \begin{cases} \|x - x'\|_2, & (x',x) \in E_H \\ c(x',x) + \varepsilon, & \text{otherwise} \end{cases} \tag{4}$$

where *c(x', x)* is the original edge cost function from Equation 2 that incorporates social cost, $E_H$ is the set of edges in the social graph *H*, and $\varepsilon$ is a small positive penalty.

The modified cost function creates a powerful exploration-exploitation trade-off. By default, the planner exploits the social graph, as traversing edges in $E_H$ incurs only the distance cost, making paths along the social graph highly attractive. However, the planner can still explore routes off the social graph by paying the original social cost plus the $\varepsilon$ penalty. The planner finds and adopts a new, superior path if one is consistently discovered, but the $\varepsilon$ penalty discourages minor, inefficient deviations from established routes. The value of $\varepsilon$ directly controls the planner's incentive to adhere to the existing social graph routes, and $\varepsilon$ must be carefully selected depending on the desired mobile robot behavior.

### 3.3 Conflict-Free Multi-Robot Trajectory Coordination

After establishing the geometric paths in Section 3.2, we compute conflict-free trajectories for a fleet of *R* mobile robots following these paths. Each robot must receive a schedule of arrival times along its planned path such that no two robots occupy the same space simultaneously, which would lead to a collision. We leverage the emergent structure of the socially constrained paths to formulate the trajectory coordination problem as a mixed-integer convex program (MICP) [34]. Unlike generic shortest-path planners, our social path planner produces routes that behave like a road network, concentrating potential collision points at a smaller set of predictable locations: intersections, opposing-direction corridors, and same direction lanes. With the social path structure, the number of collision avoidance constraints increases with the number of possible robot collisions, rather than with path length or resolution, keeping the resulting MICP more tractable than an unstructured formulation would allow. The MICP constraint generation imposes a separate computational burden on the system, and we show how the generation of the constraints can be computed in a distributed manner in Section 3.4.2. Further, the MICP formulation is flexible, supporting both sequential and simultaneous planning scenarios, suiting real-world deployments where mobile robots are dispatched on a rolling basis.

#### 3.3.1 Problem Setup: Event Waypoints and Encounter Windows

The social path planner provides each mobile robot $r$ with an ordered sequence of waypoints, $P^r = \left(p_1^r, \ldots, p_{N^r}^r\right)$. Solving for arrival times at every path waypoint scales linearly with path resolution; instead, our MICP optimizes over a sparse set of event waypoints, $W^r \subseteq P^r$. Event waypoints are selected based on their relevance to possible robot collisions and form a subset of waypoints that are sufficient to capture all relevant timing decisions. Between consecutive event waypoints, each robot is assumed to travel at constant velocity, and all remaining arrival times are recovered through interpolation.

The event waypoint set $W^r$ for mobile robot $r$ is constructed from the set of full path waypoints as illustrated in Figure 3. The start and goal waypoints are always included in $W^r$. For every other robot $s \neq r$, we identify all pairs of path segments from robots $r$ and $s$ that come within a distance $d$, the robot diameter, of each other and would lead to a collision. Contiguous conflicting segments on robot $r$'s path are merged into a single encounter window, $E(r, s)$, with entry and exit points $p_{\text{enter}}^{rs}, p_{\text{exit}}^{rs} \in P^r$. The entry and exit waypoints from every encounter window, together with the start and goal waypoints, form $W^r$. Let $K^r = |W^r|$ be the number of event waypoints and index the event waypoints as $k = 1, \ldots, K^r$ in path order. Then, the decision variables for robot $r$ are the event waypoint arrival times $t^r = \left(t_1^r, \ldots, t_{K^r}^r\right)$. Auxiliary constants used in the constraints are the arc lengths from the start to the $k$-th event waypoint and are denoted $s_k^r$. The arc length between event waypoints $k$ and $k + 1$ is $l_k^r = s_{k+1}^r - s_k^r$.

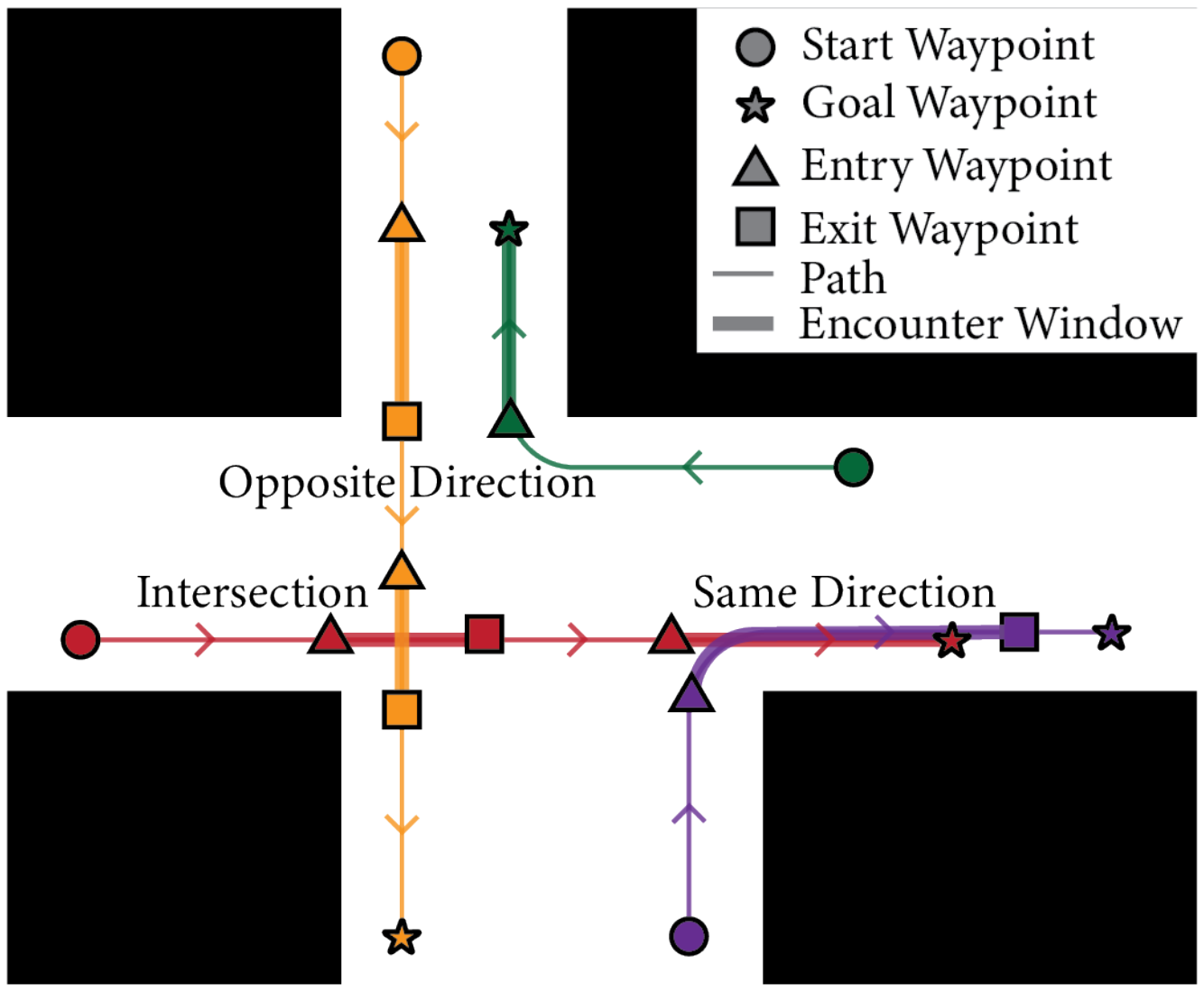

**Fig. 3.** The event waypoints and encounter windows used for generating constraints during trajectory coordination.

Because the social path planner leads to distinct and separated lanes, we classify each encounter window $E(r, s)$ into one of the following three types of encounters:

- **Intersection**: Both robots' conflict arc lengths are short (below a threshold), meaning the robots cross at a point rather than sharing a corridor. One robot must fully clear the encounter window before the other enters.
- **Opposite Direction**: The robots traverse the same corridor segment in opposing directions (detected by anti-diagonal structure in the conflicting waypoint index pairs). One robot must fully clear the corridor before the other enters.

- **Same Direction**: The robots travel the same corridor in the same direction, frequently along an identical path due to the social graph. A consistent leader-follower ordering is enforced (the leading robot must remain ahead of the following robot) throughout the entire encounter window.

### 3.3.2 MICP Constraints

The MICP optimizes the event waypoint arrival times for all mobile robots, subject to constraints along their planned social paths. The MICP contains two classes of linear constraints: kinematic constraints and collision avoidance constraints.

**Kinematic Constraints**: Each mobile robot must not exceed its maximum speed, $V^r_{\max}$, to operate safely and within physical limits. The constant velocity assumption provides a minimum traversal time between each event waypoint based upon the arc length between consecutive event waypoints:

$$t^r_{k+1} - t^r_k \geq \frac{l^r_k}{V^r_{\max}},\ k = 1, \ldots, K^r - 1 \tag{5}$$

with boundary condition $t^r_1 = 0$ for all *r*.

**Collision Avoidance Constraints**: For each encounter window, *E(r, s)*, let $k^{rs}_{\text{enter}}$ and $k^{rs}_{\text{exit}}$ denote the event waypoint indices on robot *r* corresponding to the entry and exit of the conflict zone with robot *s*. Similarly, let $k^{sr}_{\text{enter}}$ and $k^{sr}_{\text{exit}}$ denote the corresponding event waypoint indices for robot *s*. A time safety margin $\delta > 0$ enforces a temporal buffer between mobile robots reaching event waypoints that would cause a collision. The value of $\delta$ should be chosen based on the mobile robot's control fidelity and timing uncertainties. We now describe the collision constraints required under the three encounter window classifications.

For intersection and opposite direction encounters, one robot must fully exit the conflict zone before the other enters, but the order in which the robots traverse the conflict zone is not specified. This either-or constraint creates a logical disjunction, which is not directly solvable in a convex program. We encode the disjunctive constraints into the MICP using a large positive constant *M* and a single binary variable $z^{rs} \in \{0, 1\}$ (the Big-M method), a standard technique for handling logical disjunctive constraints in mixed-integer programs [35]:

$$t^r_{k^{rs}_{\text{exit}}} \leq t^s_{k^{sr}_{\text{enter}}} - \delta + Mz^{rs} \tag{6A}$$

$$t^s_{k^{sr}_{\text{exit}}} \leq t^r_{k^{rs}_{\text{enter}}} - \delta + M(1 - z^{rs}) \tag{6B}$$

The Big-M method ensures that both Equation 6A and Equation 6A hold, but only one of the constraints is active. When $z^{rs} = 0$, Equation 6B is trivially satisfied, and robot *r* exits the conflict zone before robot *s* enters. When $z^{rs} = 1$, Equation 6A is trivially satisfied, and robot *s* exits the conflict zone first. The MICP solver selects the ordering of robots to traverse the encounter window that minimizes the objective (which we will formally define in Section 3.3.3). Exactly one binary variable is required per encounter window.

For same direction encounters, a leader-follower ordering is enforced within the encounter windows. Mobile robots are allowed to be within the same encounter window simultaneously but must maintain strict ordering to prevent collisions. The leader-follower ordering is imposed at every event waypoint within the encounter window, not only at the entry and exit event waypoints associated with the encounter window. For example, if an encounter window *E(r, t)* begins or ends within *E(r, s)*, an event waypoint associated with *E(r, t)* lies within *E(r, s)* and the leader-follower ordering must also be obeyed at this intermediate event waypoint. Let $K^{rs}$ denote the ordered set of event waypoint index pairs $(k^r, k^s)$. Here, $k^r$ is the index for a robot $r$ event waypoint lying within the same direction encounter window *E(r, s)* and $k^s$ is the corresponding event waypoint index for robot $s$ at the same normalized arc length distance in the encounter. The leader-follower constraints for encounter window *E(r, s)* are again disjunctive:

$$t_{k^r}^r \le t_{k^s}^s - \delta + Mz^{rs}, \forall\ (k^r, k^s) \in K^{rs} \tag{7A}$$

$$t_{k^s}^s \le t_{k^r}^r - \delta + M(1 - z^{rs}), \forall\ (k^r, k^s) \in K^{rs} \tag{7B}$$

The constraints in Equation 7 maintain consistent robot ordering at every shared event waypoint in the encounter window, preventing a collision between mobile robots traveling in the same direction under the assumption of constant velocity between event waypoints. Note that the same binary variable $z^{rs}$ is used for all pairs in $K^{rs}$, enforcing a consistent leader-follower ordering throughout the entire encounter window.

**Special Cases**: When the start waypoint of robot $r$ is also an entry waypoint of an encounter window *E(r, s)*, the ordering must be fixed: robot $r$ must depart first since no valid ordering would otherwise exist. Similarly, if the goal waypoint of robot $r$ coincides with an exit waypoint, robot $r$ must reach the goal after any other robot that has a conflict at that location. These fixed ordering boundary conditions eliminate the need for a binary variable in the affected encounter, reducing the number of integer variables and simplifying the MICP. Furthermore, if robots $r$ and $s$ have multiple disjoint encounter windows, each window must use its own binary variable.

**Big-M Constant (*M*)**: The constant $M$ must be large enough to ensure the inactive constraint is non-binding, but not so large as to cause numerical instability in the solver. We compute $M$ dynamically as an upper bound on the total trajectory horizon:

$$M = 2\sum_r \sum_k \frac{l_k^r}{V_{\max}^r} \tag{8}$$

which yields an $M$ on the order of twice the maximum possible single robot completion time, providing a bound that avoids numerical instability from arbitrarily chosen large, fixed constants.

### 3.3.3 MICP Formulation

The complete MICP is formalized as

$$\text{minimize} \quad \sum_{r=1}^{R} t_{K^r}^r$$

$$\text{subject to} \quad t_1^r = 0 \qquad \forall\, r$$

$$t_{k+1}^{r} - t_{k}^{r} \geq \frac{l_{k}^{r}}{V_{\max}^{r}} \qquad K = 1, \ldots, K^{r} - 1, \forall\, r \tag{5}$$

$$t_{k_{\text{exit}}^{rs}}^{r} \leq t_{k_{\text{enter}}^{sr}}^{s} - \delta + Mz^{rs} \qquad \forall \text{ intersection/opposite encounters, } (r, s) \tag{6A}$$

$$t_{k_{\text{exit}}^{sr}}^{s} \leq t_{k_{\text{enter}}^{rs}}^{r} - \delta + M(1 - z^{rs}) \qquad \forall \text{ intersection/opposite encounters, } (r, s) \tag{6B}$$

$$t_{k^{r}}^{r} \leq t_{k^{s}}^{s} - \delta + Mz^{rs} \qquad \forall \text{ same direction encounters, } \forall\, (k^{r}, k^{s}) \in K^{rs} \tag{7A}$$

$$t_{k^{s}}^{s} \leq t_{k^{r}}^{r} - \delta + M(1 - z^{rs}) \qquad \forall \text{ same direction encounters, } \forall\, (k^{r}, k^{s}) \in K^{rs} \tag{7B}$$

The objective function is the sum of completion (SOC) times across all mobile robots, since $t_{K^r}^{r}$ is the arrival time to the goal waypoint for robot *r*. The solver minimizes the SOC by jointly choosing robot orderings $z^{rs}$ and event waypoint arrival times $t^r$. Alternatively, the objective could minimize the makespan ($\max_r t_{K^r}^{r}$), but we use the SOC objective to maintain linearity in the optimization problem. Thus, the problem is a mixed-integer linear program, with the objective and all constraints linear, and the variables $z^{rs}$ required to be 0 or 1. The number of binary variables equals the number of non-boundary encounter windows, so the integer programming complexity grows with fleet encounters, not with path length or resolution.

The MICP formulation above is general and accommodates three fleet coordination scenarios without changing the constraint structure; only the variables that are free or constant differ. In the sequential scenario, a new robot *r* is dispatched while existing robots are executing already known trajectories. In the constraints, robot *r*'s event waypoint arrival times are decision variables and all other robot arrival times are constant. In the simultaneous scenario, a fleet of *R* robots is dispatched at once and the event waypoint arrival times for all robots are decision variables. In the unified scenario, a set of planning robots must plan around a set of robots that are already executing fixed trajectories. In the unified case, the planning robot event waypoint arrival times are decision variables, and the already planned robot arrival times are constant. Regardless of the scenario, the constraints in Equations 5-7 are constructed identically, adapting the optimization problem to any scenario.

After the MICP is solved, the full trajectory for each mobile robot is recovered through linear interpolation of arc length between consecutive event waypoints, achieving a piecewise constant velocity motion profile. Although robot controls may not directly adhere to the constant velocity profile (due to finite acceleration/deceleration), the time safety margin $\delta$ provides a collision buffer if deviations from the constant velocity assumption are minimal.

Our MICP formulation does not guarantee feasibility under all conditions. Environments with narrow corridors or highly congested encounter windows could lead to no feasible solution. Should the MICP prove infeasible, potential mitigations include relaxing timing constraints or diverting a subset of robots away from a congested route to distribute demand across the social graph, although these are not implemented in this work. We also note that because our formulation resolves the ordering centrally and before execution, classical deadlock cannot occur at the global planning level. Deadlock could instead arise during the local planner's collision avoidance, but this concern falls outside the scope of this work.

### 3.4 Robustness and Scalability

For the proposed framework to be viable in real-world settings, we address two practical challenges: reacting to dynamic changes in the environment and reducing the computational burden of constraint generation as fleet size grows. This section details the strategies developed for each challenge.

### 3.4.1 Unmapped and Persistent Obstacles

While the global map is assumed to be static, new unmapped obstacles may be added to real-world environments, for example, repositioned furniture, a parked cart, or a delivery box. We provide a mechanism to update the global map and future global planning due to detections of new persistent static objects. This mechanism is not used for dynamic replanning during a route, however, which remains the role of the local planner. The framework accommodates map changes by leveraging temporary occupancy grid overlays and heatmaps. The process is as follows:

1) Detection and Local Invalidation: The local planner and computer vision modules on the mobile robot are responsible for detecting and classifying obstacles from onboard sensor data. When an object is detected that is static and is not accounted for in the existing map, the local planner reports the new object to the global planner. The shared social graph $H$ is not immediately modified upon receiving the new object. Instead, the global planner creates a temporary, local overlay in the occupancy grid representing the new obstacle. This staged process prevents temporary static obstacles from immediately affecting the fleet's shared social graph.
2) Global Path Updates: The global planner computes updated paths for future traversals of the affected area considering the original occupancy grid, the social graph, and the temporary overlay. The local overlay forces the planner to deviate from the social graph in the obstructed area. The cost function in Equation 4 naturally responds to the deviation, with the planner paying the social cost $\phi$ and penalty $\varepsilon$ to find a new, socially compliant detour.
3) Path Broadcasting with Context: New global paths that require a detour due to the new object are broadcast to the fleet but tagged as "temporary." These "temporary" paths are aggregated into a separate, temporary set of heatmaps. The primary social graph $H$ is built only from non-temporary paths, insulating the social graph from transient behavior. Only paths that come within a specified proximity of the obstacle should be marked as temporary.
4) Adaptation and Recovery: Our approach enables both recovery from temporary static objects and long-term adaptation for permanent static objects:
    - Recovery: When a temporary object is removed, as detected by the local planner, the global planner clears the temporary overlay for the object. Since the temporary detour paths are not included in the social graph, the mobile robot reverts to the original social graph.
    - Adaptation: If an object becomes permanent, many mobile robots will generate the same temporary detour paths around the object. If the detour count in the temporary heatmap surpasses a threshold, the system can promote the temporary heatmap to the social heatmap while removing the edges in the social graph now blocked by the obstacle, permanently updating the social graph to reflect the new obstacle.

Our strategy ensures that the system adapts to both permanent and temporary obstacles without sacrificing the long-term integrity of the social graph.

### 3.4.2 Distributed Constraint Generation for Scalable Trajectory Construction

While the trajectory coordination problem is convex, the process of identifying all potential conflicts to generate constraints can become a bottleneck. The number of pairwise collision checks grows quadratically ($O(R^2)$) with the number of robots $R$ in the simultaneous planning case, making centralized constraint generation slow for large fleets. To improve scalability, the mobile robot fleet acts as a distributed computing system, parallelizing the constraint generation process. This section details the distributed protocols for each planning scenario. We assume that communication latency is negligible compared to collision checking computation time, but we explore the role of communication latency during experiments in Section 4.4.3.

The distributed constraint checking strategies for each of the planning scenarios are as follows:

- Sequential Planning Protocol: Under sequential planning, the planning robot must check for conflicts against $R - 1$ other robots. First, the planning robot broadcasts its waypoints to the fleet. Each of the other $R - 1$ mobile robots independently computes the collision constraints with its own trajectory and transmits the conflicts to the planning robot. The planning robot gathers the constraints and solves the MICP.
- Simultaneous Planning Protocol: Under simultaneous planning for $R$ mobile robots, $\frac{R(R-1)}{2}$ pairwise collision checks are distributed. Each mobile robot broadcasts its path and uses a deterministic algorithm to determine which constraint checks to perform. The robots use a circular ordering scheme illustrated in Figure 4, where each robot computes constraints for approximately $n = \frac{R-1}{2}$ robot pairs. If $R$ is odd, every mobile robot performs $n$ robot pair checks. If $R$ is even, half of the mobile robots perform $n + 1$ robot pair checks, and the other half perform $n$ robot pair checks. After checking for possible collisions, each mobile robot broadcasts the discovered encounter windows, and each robot can locally build and solve the identical, full MICP.
- Unified Planning Protocol: The distributed approach for the unified scenario is a combination of the sequential and simultaneous scenarios. The currently planning robots are responsible for distributing collision checks using the procedure from simultaneous planning. Meanwhile, the already planned robots generate the sequential collision constraints of their own trajectory with each of the currently planning robot paths.

Distributing the constraint generation reduces the total computational time required to formulate the MICP problem by leveraging the entire fleet's computing power. The resulting convex program, however, is still solved centrally. We show in Section 4.3.2 that, across the fleet sizes we evaluate, our approach maintains a higher success rate and shorter solve time than existing approaches that do not consider the social structure.

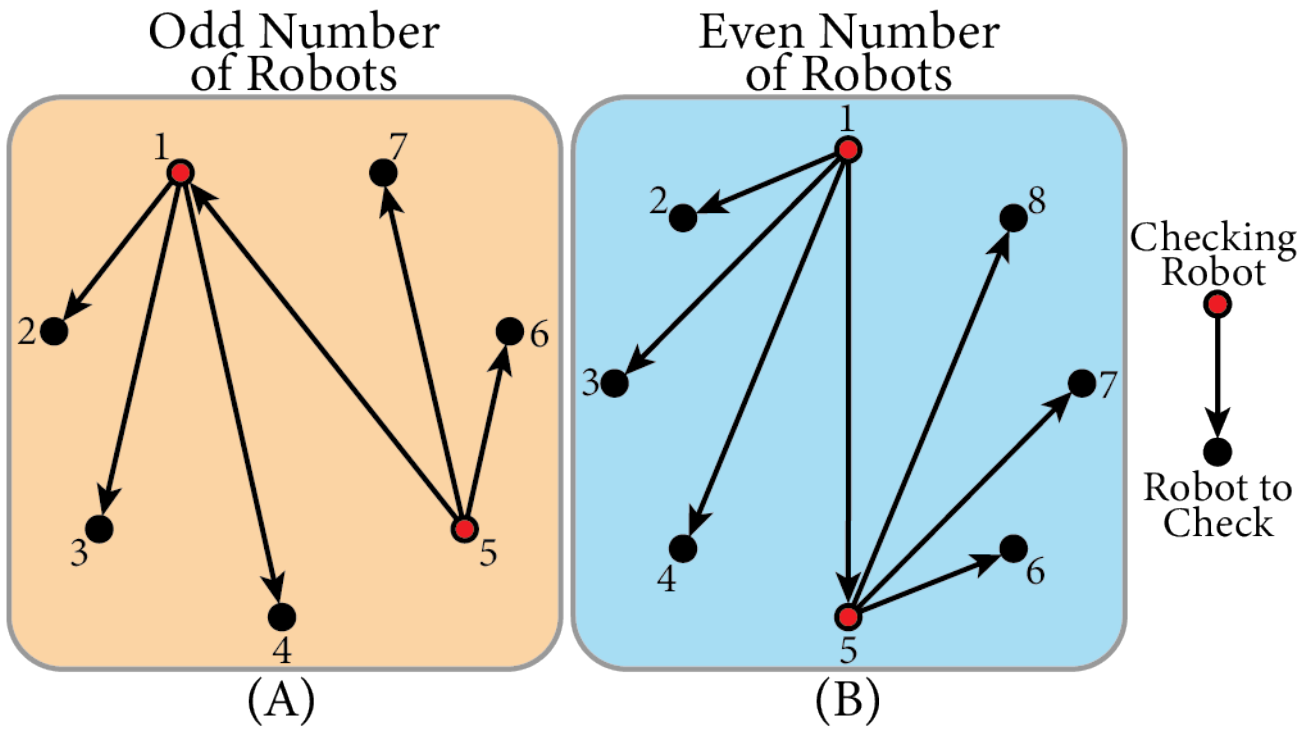

**Fig. 4.** Distributed collision checking assignments for robots 1 and 5 in the case of (A) an odd number of robots and (B) an even number of robots.

## 4. Results

To validate the performance and characteristics of our proposed framework, we conduct a series of experiments on simulated robot environments and demonstrate the framework on hardware. First, we analyze the social path planner's ability to generate paths that align with the desired social characteristics. Then, we demonstrate the utility of the social graph, assessing the social graph's impact on path consistency and planning efficiency. Finally, we evaluate the performance and scalability of the MICP trajectory generator for multi-robot coordination. The experiments are conducted in the three environments shown in Figure 5, which include two layouts with idealized hallway and room structures and one environment generated by mapping the second floor of the Y2E2 building at Stanford University. Environments 1 and 2 are used entirely in simulation, while Environment 3 is used primarily for the hardware experiments in Section 4.4. The parameters used for these experiments are listed in Table 1.

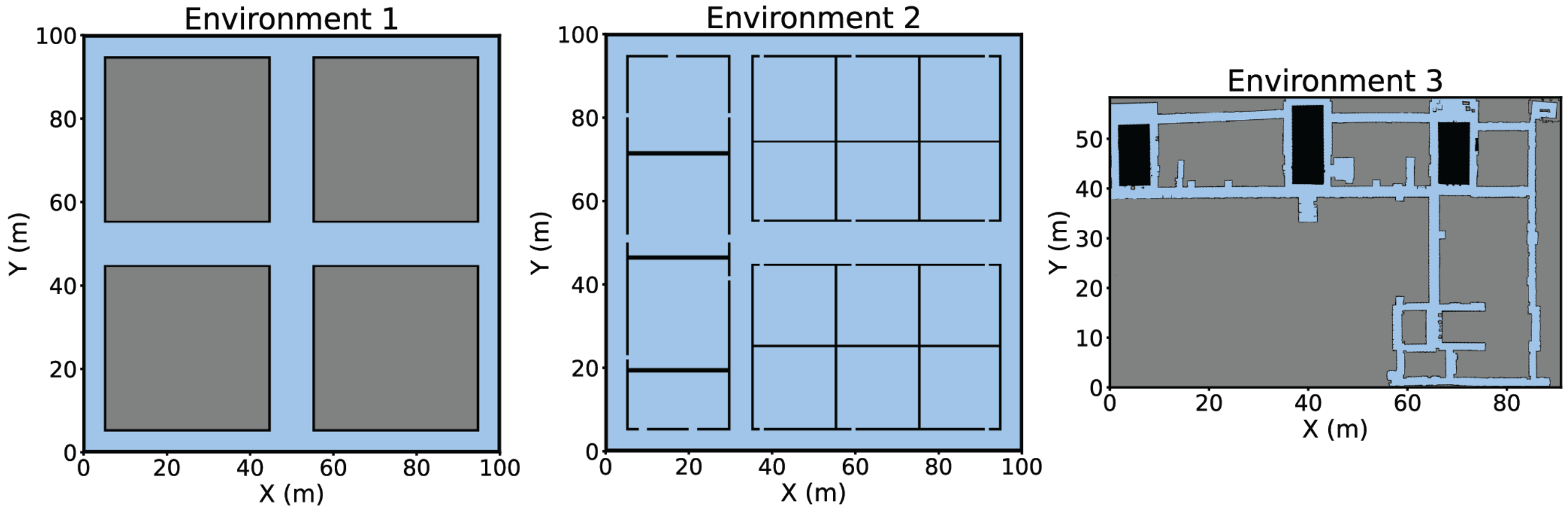

**Fig. 5.** The test environments for the path planner and trajectory controller. Environments 1 and 2 are artificially created environments, while Environment 3 is a real map from the second floor of the Y2E2 building at Stanford University.

**Table 1.** The parameters selected for the experiments.

| Parameter | Value |
|---|---|
| $w_{\text{right}}$ | 1.0 |
| $w_{\text{const}}$ | 2.0 |
| $w_{\text{buf}}$ | 5.0 |
| $d_{\text{buf}}$ | 4.0 m |
| $\varepsilon$ | 0.01 |
| $V_{\text{max}}$ | 0.7 m/s |
| $\delta$ | 1 s |

### 4.1 Social Planner Performance and Compliance

We first evaluate the social planner's ability to generate paths that balance efficiency with social compliance. To quantify the social planner's success, we compare the planner against the vanilla A* (shortest path) and RRT* (sampling-based planner) algorithms. The evaluation consists of 500 trials for each environment, with each trial using a random start and goal location. We measure three metrics: total path length (efficiency), average distance to the right-hand wall (social compliance), and computation time.

The results, presented in Table 2, reveal the expected trade-offs. The social planner produces plans that are, on average, longer than the optimal paths found by A*. The increase in path length is a direct result of enforcing social rules, which often preclude taking the shortest possible route. In exchange for the modest increase in path length, the social planner demonstrates improved social compliance, maintaining a much smaller and more consistent average distance to the right-hand wall. The results also highlight the larger computation cost of the social planner, due to the more complex cost function. However, we show in Section 4.2 that the planning speed can be improved dramatically by developing and using a social graph. Additionally, while not used in this experiment, the social cost can be precomputed for each valid movement, eliminating the cost of computing the social cost function during planning.

**Table 2.** Evaluation of the social planner as compared to baseline planners A* and RRT*. Each comparison uses the same 500 randomly sampled start and goal locations for each method.

| Algorithm | Evaluation Metric: Mean Path Length (m) | Mean Right Distance (m) | Mean Computation Time (sec) |
|---|---|---|---|
| | **Environment 1** | | |
| A* | 64.3 ± 33.3 | 8.31 ± 8.49 | 0.348 ± 0.697 |
| RRT* | 68.1 ± 37.0 | 27.3 ± 38.1 | 0.764 ± 1.126 |
| Social Planner | 74.0 ± 37.4 | 2.64 ± 5.95 | 20.7 ± 18.0 |
| | **Environment 2** | | |
| A* | 77.5 ± 36.6 | 12.6 ± 14.5 | 6.40 ± 8.39 |
| RRT* | 89.4 ± 48.2 | 24.7 ± 26.2 | 31.0 ± 35.7 |
| Social Planner | 94.1 ± 44.4 | 9.69 ± 15.2 | 56.8 ± 51.7 |
| | **Environment 3** | | |
| A* | 46.6 ± 26.0 | 2.43 ± 2.21 | 3.78 ± 5.19 |
| RRT* | 49.7 ± 28.1 | 10.1 ± 20.0 | 6.36 ± 12.18 |
| Social Planner | 52.9 ± 29.8 | 1.17 ± 1.12 | 30.6 ± 22.6 |

Figure 6 provides a qualitative illustration of the different planners on Environment 1, showcasing typical paths generated by each planner. The A* planner optimizes solely for path distance, producing paths that are short but often moving along the left-hand side or the middle of hallways. RRT* generates paths that are both longer than necessary and poorly compliant with social conventions, showing a high degree of variance due to the algorithm's stochastic nature. In contrast, our social planner adheres to the social rules, traveling only along the right-hand side of the hallways while taking only slightly longer paths.

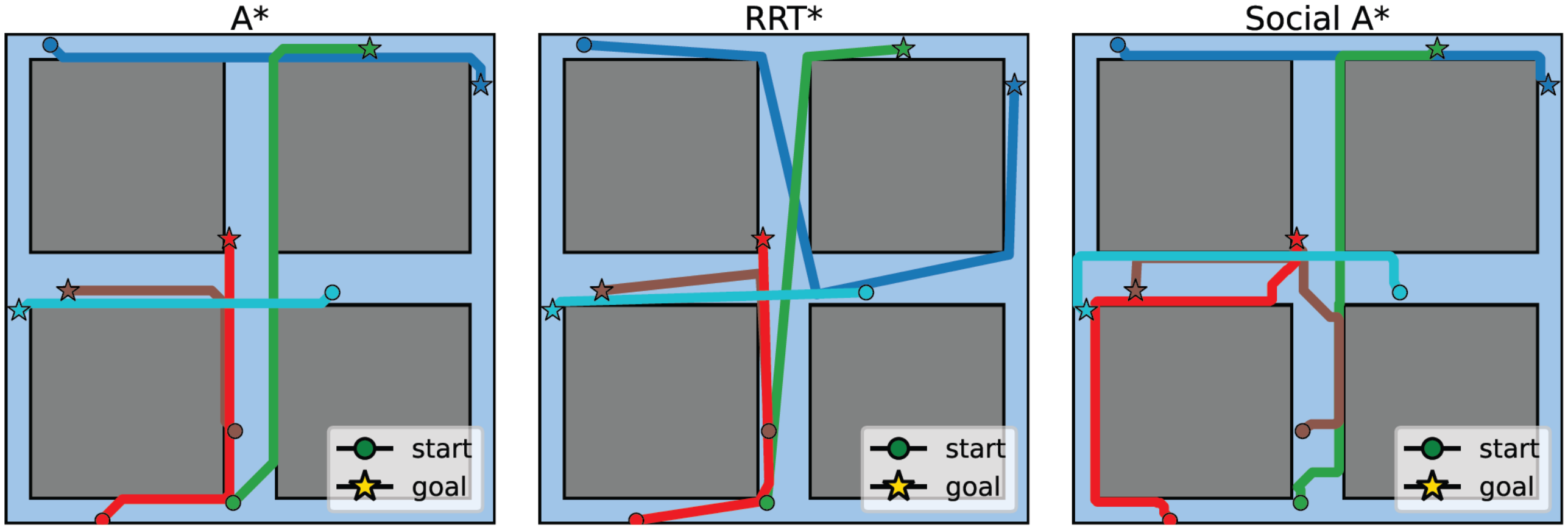

**Fig. 6.** Examples of paths planned by A*, RRT*, and the social planner on Environment 1.

### 4.2 Social Graph Analysis

Next, we analyze the performance of the social graph, which is designed to enforce fleet-wide path consistency and lower the computation cost of social planning. First, we demonstrate the formation and convergence of the social graph, as depicted visually in Figure 7 for Environment 1. Starting with an empty graph, we use the social planner to plan paths between random points, aggregating these paths into a heatmap. As more paths are planned, distinct, highway-like lanes emerge, particularly along the walls, demonstrating that the graph successfully captures the desired social conventions and produces socially consistent routes across the fleet.

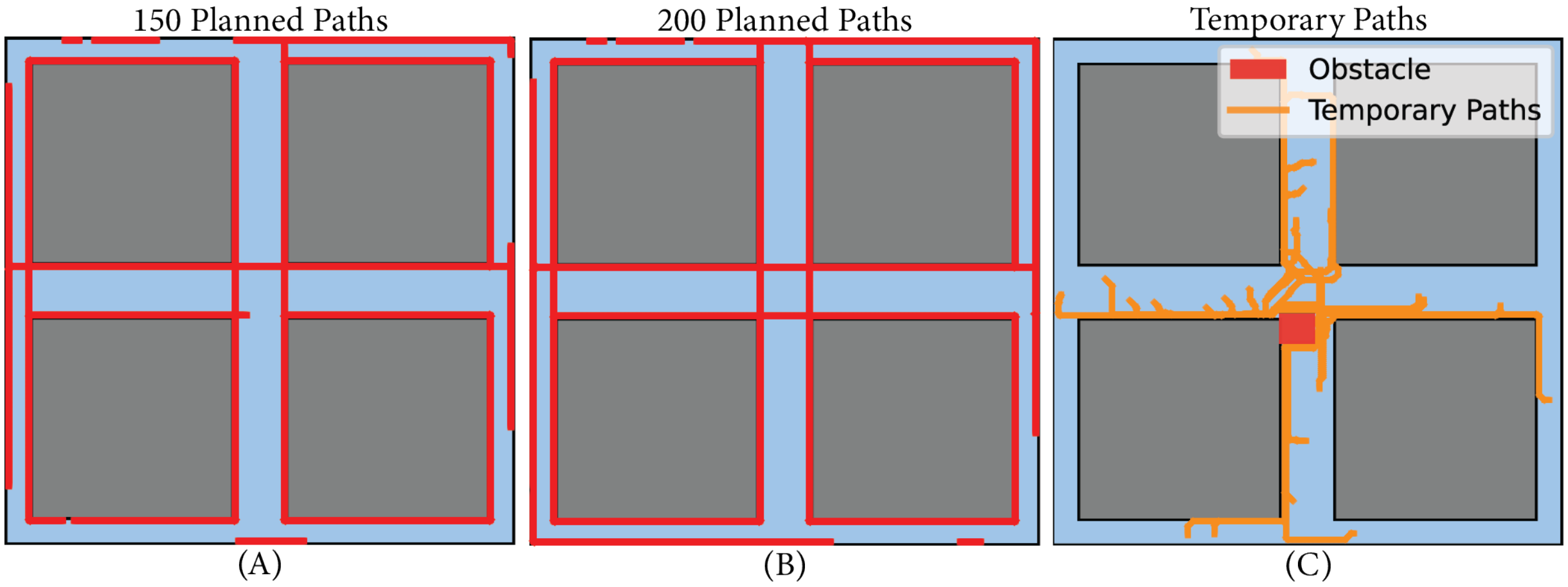

**Fig. 7.** The social graph of Environment 1 after (A) 150 planned paths, (B) 250 planned paths, and (C) with temporary paths due to a temporary obstacle.

We demonstrate the social graph under environmental changes as described in Section 3.4.1. Figure 7C shows the social graph's behavior when an object blocks a global route. New paths are routed around the new obstacle, but any path that is affected is labeled "temporary" and is not integrated into the original heatmap of the primary social graph. The temporary heatmap ensures that transient obstacles do not permanently degrade the established social graph once the obstacle is removed. Should the object become permanent, the temporary paths shown in Figure 7C can be added to the original heatmap to generate an update of the social graph for the new geometry.

To quantify route consistency directly, we compute the Shannon entropy [36] of travel direction at each occupancy grid cell. Each planner plans a path between the same 100 randomly sampled start and goal locations, and every path is converted to a sequence of occupancy grid cells. For each grid cell visited by at least three paths, we build an eight-bin histogram of the direction taken to the next cell. The Shannon entropy is computed for each cell, where a value of zero indicates all paths exit the cell in the same direction, and a value of three indicates exits are uniformly distributed across all eight directions. Cells that are rarely visited (less than three visits) are excluded from the statistics. The unweighted mean entropy for all qualifying cells for each planner is listed in Table 3.

**Table 3.** Mean Shannon entropy of per-cell exit directions for A*, RRT*, and the social planner with and without the social graph, all evaluated on identical start-goal pairs. Lower entropy indicates that independently planned paths more consistently exit each grid cell in the same direction.

| | **Mean Entropy (↓)** |
|---|---|

| Algorithm | Environment 1 | Environment 2 |
|---|---|---|
| A* | 0.654 ± 0.431 | 0.807 ± 0.345 |
| RRT* | 0.855 ± 0.493 | 0.919 ± 0.483 |
| Social A* without Social Graph | 0.054 ± 0.213 | 0.115 ± 0.300 |
| Social A* with Social Graph | 0.039 ± 0.179 | 0.049 ± 0.197 |

As shown in Table 3, the social planner, both with and without the social graph, achieves significantly lower mean entropy than A* and RRT*. The socially compliant paths converge on consistent routes far more often than unstructured planners, even without the social graph. Adding the social graph further reduces entropy in both environments, for example a 57% reduction in Environment 2. Fleet-wide route consistency therefore depends on the social graph and is not fully achieved by socially compliant paths alone.

Last, we quantify the computational benefit of planning while using the social graph. Throughout the social graph formation process, we record the time required to plan 100 paths between the same random start and goal points using the biased cost function from Equation 4. The result, plotted in Figure 8, shows that the social graph significantly reduces the social path planning time, decreasing the computation time to less than that of the A* baseline for Environments 1 and 2. The heavily biased cost function constrains the modified A* search space to the social graph's established paths, significantly reducing the required number of explored nodes. Thus, biasing the planner towards the social graph substantially reduces planning time while preserving path consistency.

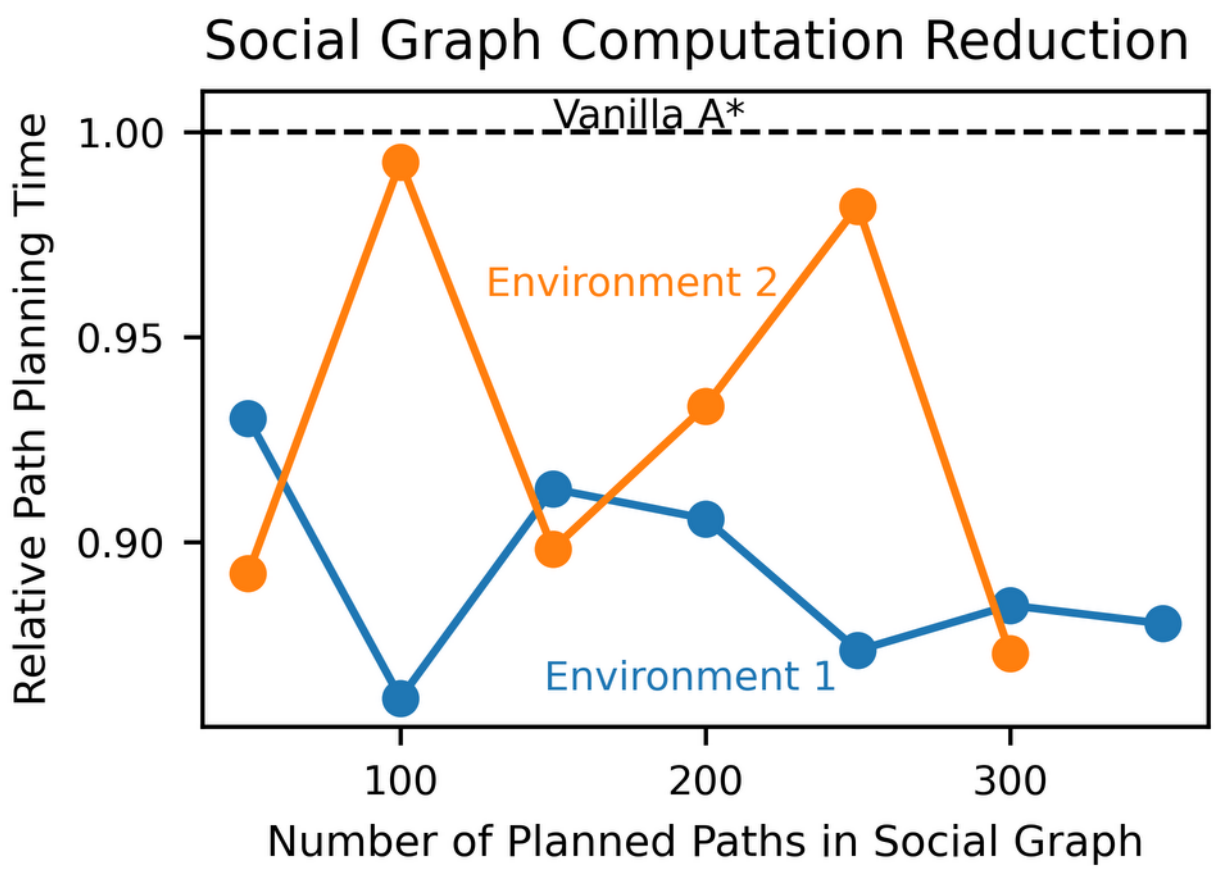

**Fig. 8.** The relative social path planning time when using the modified cost function with the social graph, as compared to vanilla A*.

### 4.3 Multi-Robot Trajectory Coordination

We evaluate the performance and scalability of the trajectory controller subject to planned social paths. The evaluation targets human-centric environments such as offices, hospitals, and airports, where fleet size is bound by both the physical space and the social conventions the robots should follow—massive fleets of robots sharing a hallway with pedestrians is atypical of a human-centric environment and more characteristic of a warehouse or logistics-style swarm robotics application with limited human presence. We therefore evaluate performance and scalability at moderate fleet sizes that are representative of our target domain. First, we provide a qualitative example to demonstrate the effectiveness of the trajectory coordination. We then compare our approach against existing multi-robot coordination methods, before conducting a quantitative analysis of computational scalability for larger fleet sizes.

### 4.3.1 Collision Avoidance

To validate the trajectory controller's ability to generate collision-free trajectories, we design a scenario in Environment 1 where four mobile robots with various maximum speeds have planned paths that intersect or overlap. The trajectory controller solves the simultaneous optimization problem described in Section 3.3.3, with Figure 9 showing the robots as they approach the intersection. The controller successfully computes arrival times for each mobile robot without collisions and optimizes for the total time taken by the mobile robots. Notice that the faster robots allow the slower gold robot to pass through the intersection first, so it can follow its path without slowing. Meanwhile, the two robots with the same direction encounter window maintain their ordering along the entire segment. Figure 9 shows the utility of the social graph, with planned paths only overlapping at intersections or when moving in the same direction along the same path, simplifying the potential collision locations.

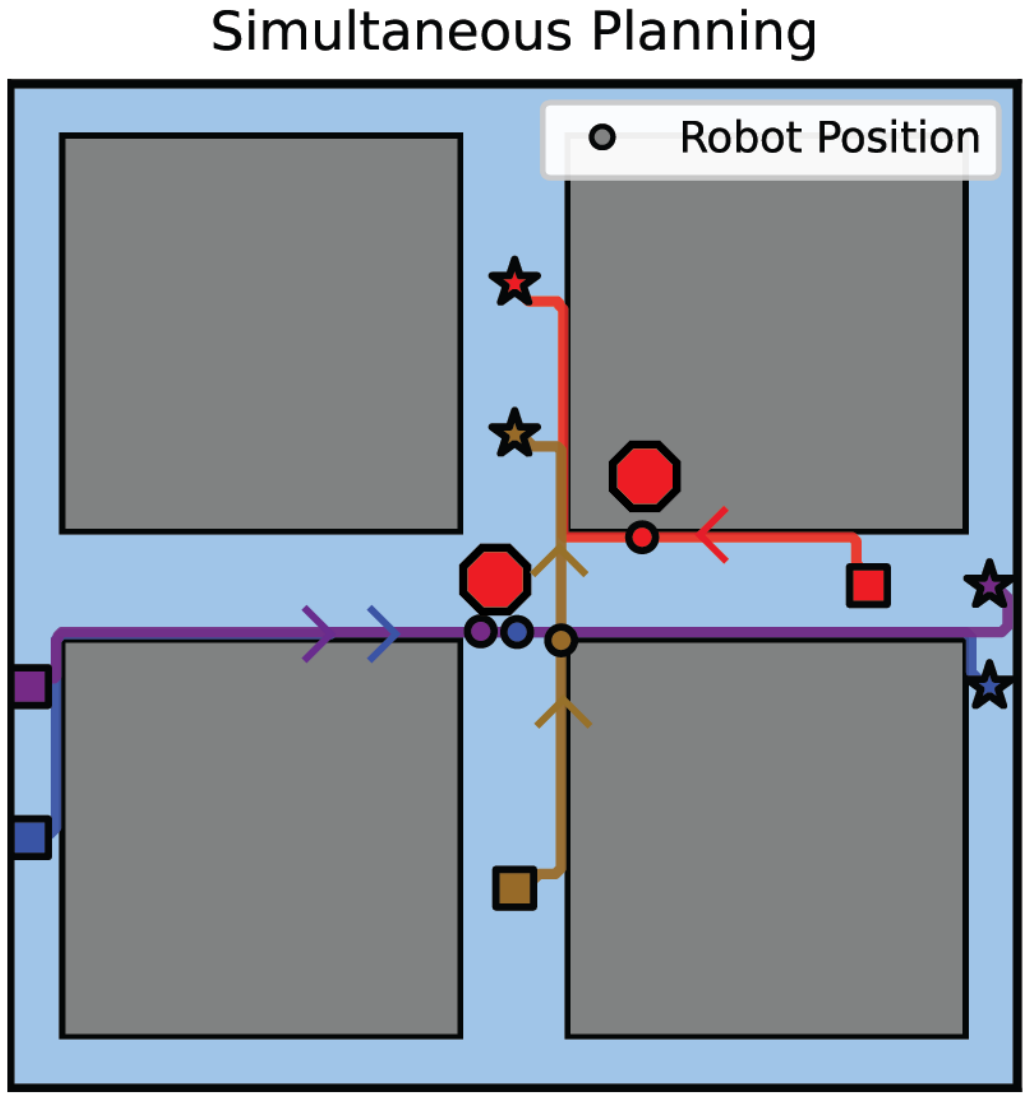

**Fig. 9.** An example of optimized trajectories under simultaneous planning for four mobile robots. The other mobile robots allow the slowest robot (gold robot) to pass through the intersection first. The blue and purple robots maintain a consistent ordering throughout their entire shared path segment.

### 4.3.2 Comparison to Baseline Approaches

We evaluate our trajectory coordination approach on Environment 1 using the MICP against the conflict-based search (CBS) [16] method and sequential priority planning (PP) [17] method. The CBS method first plans each path individually and then resolves resulting collisions with a higher-level search algorithm. The PP method assigns higher priority to mobile robots with longer path lengths and then plans each path sequentially in order of priority. We evaluate each method across varying numbers of mobile robots and provide a summary of the success rate, makespan time, and computation time of the solver. We limit each solver to 120 seconds of solve time and otherwise consider the attempt a failure. For CBS and PP, the map resolution is downsampled by a factor of 4 to reduce the computational load. The results are summarized in Figure 10, averaging over 25 total trials for each method. First, notice that our approach has a higher success rate than the other approaches while maintaining a shorter solve time, even as the number of mobile robots grows. The feasibility and short solve times are a direct consequence of our social planner that produces paths that rarely conflict. Our MICP requires a single binary variable for each encounter window, keeping the number of variables small. Note that the solve time only includes the MICP solver time and excludes both the social path planning time and constraint generation time. However, as we have shown, the social path planning time remains fast due to the social graph, and as we will show in Section 4.3.3, the distributed constraint generation results in improved constraint generation time. The MICP solver is the only centralized component of our system, and as shown in Figure 10, the MICP solve time remains small across the fleet sizes we test, on the order of one second. In comparison, techniques such as CBS and PP are entirely centralized, resulting in large computation time as the number of robots increases. Our approach's runtime also does not increase with path resolution, in contrast to CBS and PP, which require a map downsampling factor to obtain reasonable solve times. Last, our MICP solution produces a competitive makespan compared to the CBS and PP methods, even though our social paths do not necessarily follow the shortest path routes. The social paths following the social graph allow robots to pipeline along the same routes, leading to reduced potential collision points and competitive makespans.

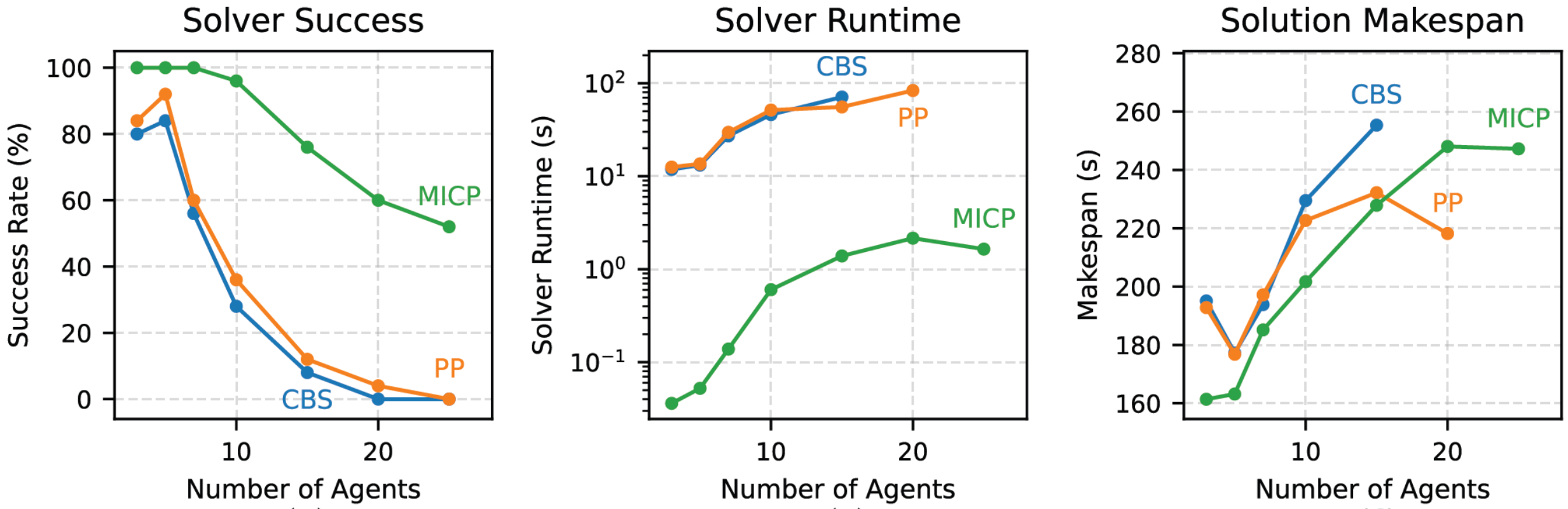

(A) (B) (C)

**Fig. 10.** The (A) solver success rate, (B) solver runtime, and (C) solution makespan for our MICP approach as compared to the CBS and PP methods as evaluated on Environment 1.

The success of our proposed trajectory coordination scheme can be attributed to the social paths. In Figure 11, we show the average number of path segments that include overlap between agents in Environment 1, averaged over 100 random scenarios for each agent count. In the overlapping segment count, we omit path segments that overlap but travel in the same direction, since travel along the same routes is desired behavior (mobile robots can pipeline along these segments and the trajectory coordinator must only solve for the relative order of the robots along these paths). As shown in Figure 11, our social planner results in fewer overlapping segments compared to A* and RRT*. With paths from our approach falling mostly along the social graph shown in Figure 7, potential collisions occur primarily at intersections, reducing the possible collision scenarios and simplifying the trajectory coordination problem. Meanwhile, for methods such as CBS and PP which rely on A*, the larger number of potential collisions leads to increased complexity in finding feasible solutions.

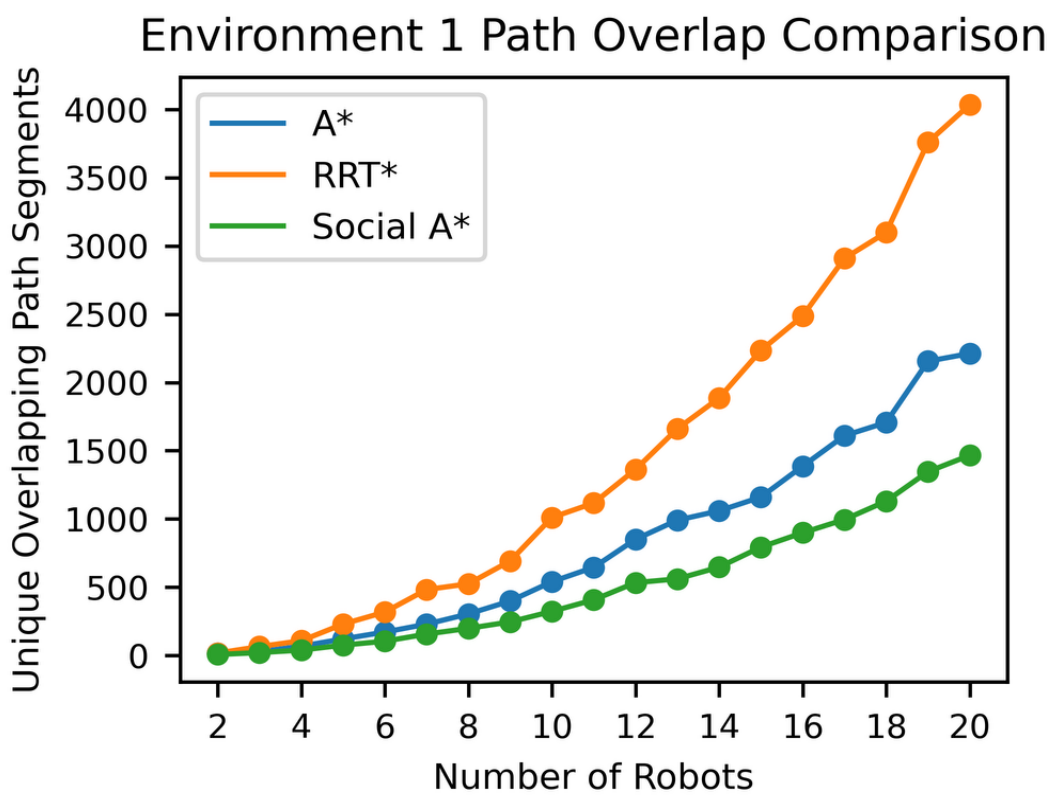

**Fig. 11.** Comparison of overlapping path segments when using the different path solvers. Overlapping segments are excluded when they travel in the same direction since social behavior encourages movement in the same lanes.

### 4.3.3 Scalability of Trajectory Optimization Constraint Generation

We analyze the scalability of the trajectory controller using the distributed constraint generation approach. Computational performance is evaluated under the most demanding scenario, simultaneous planning, which requires $O(R^2)$ collision checks to generate the optimization constraints for $R$ mobile robots. We compare a centralized constraint generation approach, where a single robot performs all collision checks, and a distributed approach described in Section 3.4.2 where collision checks are distributed across the $R$ mobile robots.

We simulate fleets of increasing sizes in Environment 2, with each robot assigned random start and goal positions. Figure 12 shows the computation time required for generating the constraints for MICP for both the centralized and distributed approach. As expected, the centralized approach computation grows quadratically since a single robot performs all $O(R^2)$ collision checks. Meanwhile, the distributed approach has a linear computation time growth, representing the computational savings from distributing the workload. The distributed framework allows the generation of the convex program to scale efficiently as the size of mobile robot fleets increases. The MICP must still be solved centrally, but the constraint generation time can be much larger than the MICP solve time shown in Figure 10B, and so the distributed approach meaningfully reduces the total problem formation and solve time.

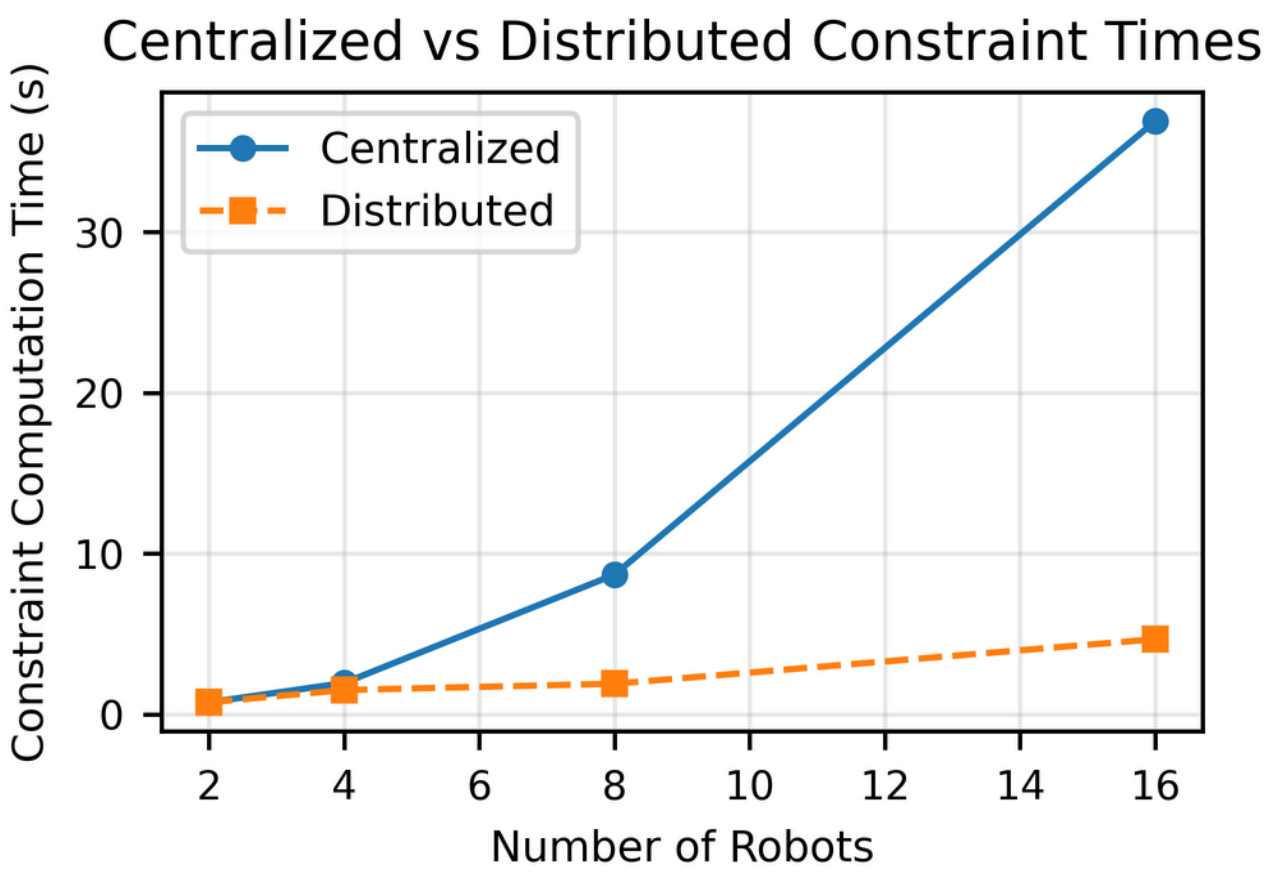

**Fig. 12.** Constraint generation computation time when using a distributed and centralized strategy.

### 4.4 Hardware Experiments and Results

We perform experiments on mobile robots in an academic office building setting and provide the results and discussion in this section. First, the experimental setup and mobile robots are introduced, and then the results are provided for common scenarios: head-on hallway traffic and intersection crossings. The implementation of our proposed path planner and trajectory coordinator, as well as supporting videos of these experiments, are available at https://github.com/satomm1/path_planning.

#### 4.4.1 Experimental Setup

The experiments are performed on custom-built circular differential drive mobile robots shown in Figure 13A. The mobile robots contain a 2D LiDAR sensor and an RGB-D camera, and have a maximum velocity of 0.7 m/s. The onboard computer is an NVIDIA Jetson Orin Nano 8 GB Developer Kit. Communication among mobile robots is performed using the Data Distribution Service (DDS) [37], a Wi-Fi-based publish/subscribe communications protocol. Software on the mobile robots is facilitated using the Noetic version of the Robot Operating System (ROS) [38].

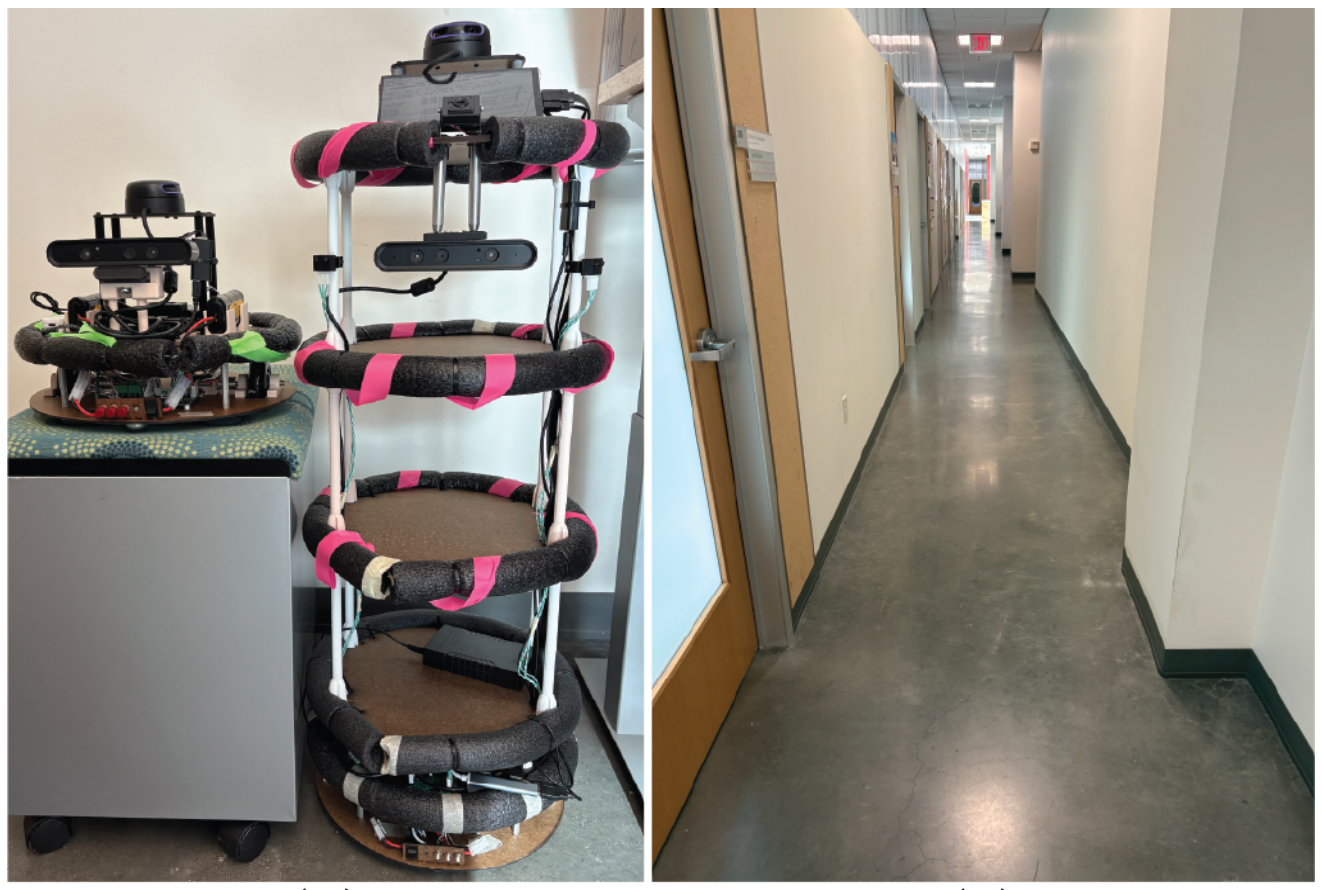

(A) (B)

**Fig. 13.** The (A) mobile robots and (B) typical building hallway used for the hardware experiments.

The experiments are performed on the second floor of the Y2E2 building at Stanford University, with the occupancy grid map shown as Environment 3 in Figure 5. The map resolution is 0.05 m per occupancy grid point, and the typical hallway width is approximately 1.5 meters, with a typical hallway of the building shown in Figure 13B. The occupancy grid is produced using the GMapping SLAM algorithm [39], and localization is performed within the map using Adaptive Monte Carlo Localization [40], both of which use a particle filter approach. Local control of the mobile robot is performed with a differential flatness controller. The MICP solver uses an embedded conic solver with a branch-and-bound procedure [41].

#### 4.4.2 Head-On Traffic in Hallways and Intersection Crossings

First, we demonstrate our approach with robots traveling in opposite directions down the same hallway. Our social rules result in the robots staying to the right-hand side of the hallway, naturally avoiding each other and bypassing the need for trajectory coordination. Consider Figure 14, which shows two different hallway scenarios featuring two mobile robots. In the upper case, the social plans lead to no collision if the mobile robots start at the same time, with both mobile robots completing their trajectory in the minimal required time. However, when using an A* path, the paths collide along most of the length of the trajectory, and the right robot must wait until the left robot completes its trajectory, resulting in a slower makespan.

Meanwhile, in the lower case, the social planner again allows both mobile robots to complete their trajectories immediately without any collision. With A*, the paths intersect throughout, and without a robot-aware local planner, the trajectory coordination is infeasible.

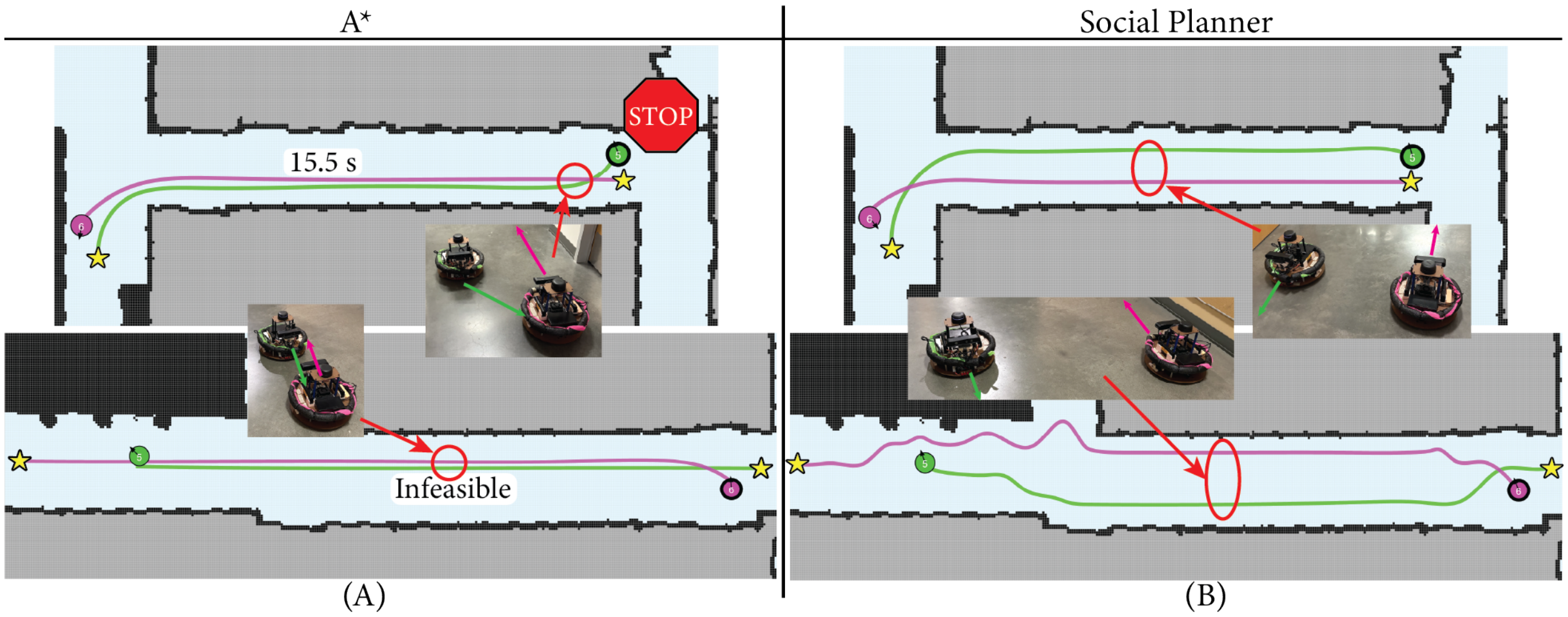

**Fig. 14.** Comparison of two mobile robots approaching each other head-on in a hallway using the (A) A* planner and (B) social planner. The social planner leads to no conflict, resulting in shorter robot navigation times.

Next, we demonstrate the framework's ability to negotiate at intersections. As shown in Figure 15, the mobile robots navigate an intersection following routes determined by the social planner with trajectories generated by our MICP formulations successfully managing the intersection scenario. In both examples, the intersection encounter window constraints dictate that only one mobile robot is allowed in the intersection at a time, while other robots must wait outside the intersection to provide a clear path. These demonstrations illustrate that our social path planning and trajectory coordination approach is effective on hardware, limiting the possible collision zones and successfully assigning event waypoint arrival times.

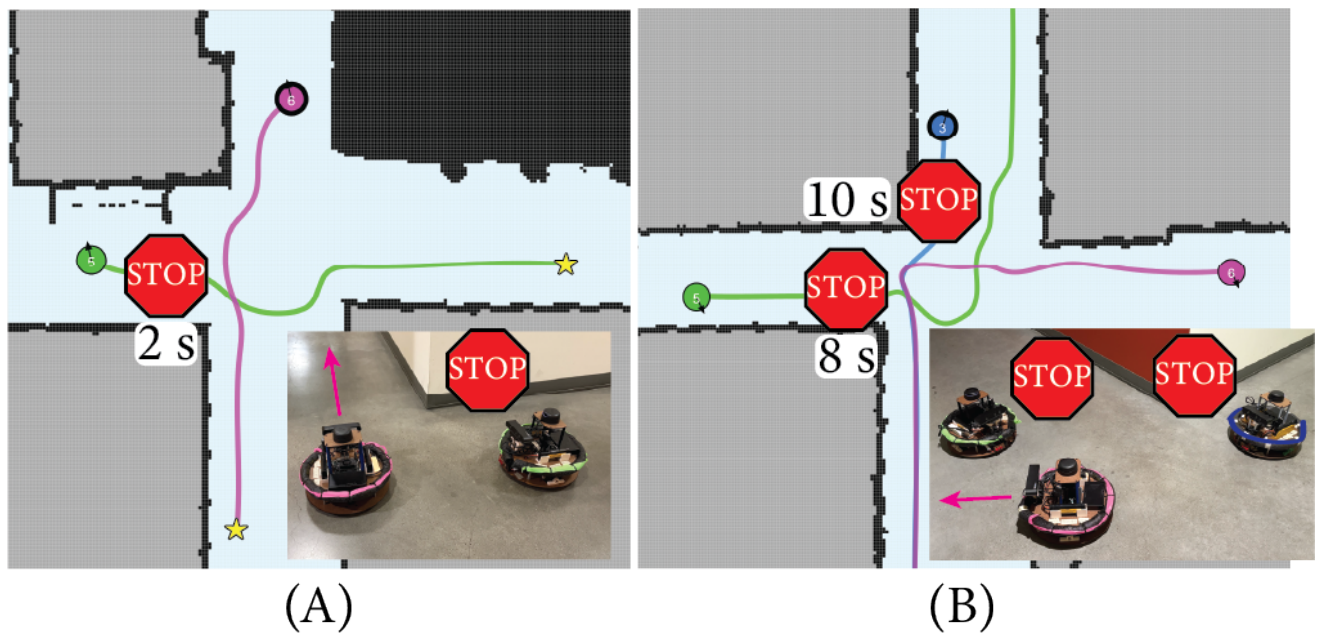

**Fig. 15.** Demonstration of (A) two and (B) three mobile robots negotiating at an intersection using the social path planner and trajectory coordination.

### 4.4.3 Communication Latency and Distributed Computing

We record communication latencies of less than 100 ms when communicating paths and collision constraints between mobile robots using the Wi-Fi based DDS communication system. Meanwhile, generating collision constraints for a single pair of robots is typically greater than 200 ms depending on the path lengths. Thus, communication latency is small enough to provide benefit for even a small number of mobile robots. Table 4 provides the average relative speedup in constraint generation when using a distributed approach instead of a centralized approach on our robot hardware. We see an improvement of at least 50%, with the relative speedup increasing as the fleet size grows. The improvement follows directly from the centralized method's quadratic scaling and the distributed method's linear scaling. These hardware results confirm that low communication latency allows distributed constraint generation to meaningfully reduce total trajectory coordination time, consistent with the results in Section 4.3.3.

**Table 4.** Average improvement in the constraint generation time on robot hardware when using the distributed approach as compared to the centralized approach.

| Number of Robots | Mean Improvement |
|---|---|
| 3 | 55.1% ± 13.2% |
| 4 | 67.2% ± 6.9% |
| 5 | 78.4% ± 1.8% |

#### 4.4.4 Local Planner Considerations

Our framework provides a global path and trajectory for multiple robots. However, due to localization noise, actuation error, and dynamic objects in the environment, the framework must be supplemented with a local (reactive) path planner. To maintain social compliance, the local planner should also be human-aware. To fully maintain the multi-robot advantages of our proposed framework, the local planner should be tightly integrated with the global planner, such that the robot returns to the global path as soon as possible to maintain the validity of the collision-free trajectories. Additionally, the local planner should account for the planned trajectories of the other mobile robots when planning local trajectories. If a mobile robot becomes significantly delayed or the planned path becomes unreachable due to a new obstacle, our framework accommodates this scenario through the sequential path planning option. The delayed or blocked mobile robot can perform a sequential path planning optimization for its remaining path, reintegrating with the existing mobile robot trajectories. If mobile robot failure affects a subset of the fleet or the entire fleet, the mobile robots can perform a unified or simultaneous global replan.

## 5. Discussion and Conclusion

The integration of autonomous mobile robots into human-centric environments depends on the mobile robots' ability to navigate safely, efficiently, and in a manner that is both socially compliant and consistent. While many existing navigation systems address local planning and collision avoidance, they do not generate globally consistent and socially acceptable fleet-wide behavior. This paper addresses the global social path planning gap by introducing a system that integrates global, socially aware path planning, fleet-wide route consolidation, and multi-robot trajectory coordination.

Human social norms are first embedded directly into the cost function of a modified A* planner, generating socially compliant paths for each robot. To address the computational overhead and enforce fleet-wide consistency, we introduce a social graph that consolidates established, socially compliant routes across the fleet, enforcing social consistency and reducing future planning effort. The socially consistent route structure leads to a tractable multi-robot trajectory coordination formulation: because robots are biased toward a shared set of social routes, the possible conflicts are reduced to a smaller set of well-defined encounters, which we exploit to formulate a mixed-integer convex program. Constraint generation for the program can be distributed across the fleet, although the resulting optimization is still performed centrally. The trajectory coordination formulation is evaluated at moderate fleet sizes in simulation and on a small hardware fleet. While social cost embedding, route reuse, and multi-robot trajectory optimization have each been studied independently in prior work, we combine these aspects into a single system for global, socially consistent multi-robot planning.

There are several limitations of the current framework worth noting. First, the social cost function used in this work is purposely simple, encoding only basic macroscopic conventions such as right-hand side travel. The social cost function does not address social factors including, but not limited to, pedestrian interactions, group behaviors, or cultural variation in navigation norms. Thus, the social cost function used here should not be interpreted as a comprehensive model of human social navigation, but rather an example of how a social cost function can be integrated into the framework. Second, our approach is best suited to environments with hallway-like structure or with clear geometric features to organize routes around; wide, open spaces without structure (atria, plazas, open floor plans) may not separate socially compliant paths into distinct, consistent lanes, which our trajectory coordination formulation relies upon to reduce the set of possible collisions. Third, our approach targets human-centric environments where robot density is moderate and is not designed or evaluated for swarm-scale deployments involving high robot density. Fourth, our trajectory coordination formulation does not provide a formal feasibility guarantee under arbitrarily congested conditions. Scenarios with narrow corridors, high encounter window density, or tight timing constraints could result in an infeasible optimization problem, and we do not provide an automated fallback in case of infeasibility in this work. Last, while the broader literature motivating this work is concerned with human comfort and acceptance, we do not directly evaluate how our robots' socially consistent paths affect human perception of comfort or acceptance. Instead, our route consistency analysis (Section 4.2) offers a proxy for the repeatability of robot behavior (one of several factors that may influence human comfort), rather than a validated measure of human-perceived comfort.

In addition to the limitations above, we identify several other directions for future work. First, transitioning from engineered global map social costs to learning behaviors directly from observed human motion data could allow mobile robots to adapt to the unique conditions of a specific environment. Second, while Section 4.4.4 outlines the local planner requirements for full deployment, integrating and validating a responsive local planner with our framework remains an important direction for future work. Finally, future work can explore how to adapt the social graph over large time scales, allowing the social graph to evolve in response to changes in human traffic patterns or building layout.

In conclusion, this work demonstrates that by enforcing social compliance and consistency at the global planning level, the complex problem of multi-robot collision avoidance becomes more tractable. Our system-level approach enables efficient path planning and coordinated trajectory generation for multi-robot fleets, all while following human social norms. Our work is an important step toward creating autonomous mobile robot systems that not only share spaces with humans, but respect human social conventions, ultimately enabling increased uptake and safer coexistence with humans.

**Acknowledgments**

**Funding:** This material is based upon work supported by the National Science Foundation (NSF) Graduate Research Fellowship Program [Grant No. DGE-2146755]. This research is also partially supported by the Stanford Center for Sustainable Development and Global Competitiveness (SDGC) and the Center for Advanced Urban Systems (CAUS) at Korea Advanced Institute of Science and Technology (KAIST). Any opinions, findings, and conclusions or recommendations expressed in this material are those of the authors and do not necessarily reflect the views of NSF, SDGC, or KAIST-CAUS.